\documentclass[10pt,twocolumn,letterpaper]{article}

\usepackage{cvpr}              % To produce the CAMERA-READY version
\usepackage{graphicx}
\usepackage{subcaption}
\graphicspath{{figs/}}
\usepackage{multirow}

\usepackage{amssymb}
\usepackage{booktabs}
\usepackage{tabularx}
\usepackage{enumitem}
\usepackage{svg}
\usepackage{tikz}
\usepackage{multibib}
\newcites{supp}{References (Supplementary)}
\usepackage{float}
\usepackage{caption}

\definecolor{cvprblue}{rgb}{0.21,0.49,0.74}
\usepackage[pagebackref,breaklinks,colorlinks,allcolors=cvprblue]{hyperref}
\usepackage[accsupp]{axessibility}

\def\paperID{} % *** Enter the Paper ID here
\def\confName{CVPR}
\def\confYear{2026}

\title{Cross-Modal-Domain Generalization Through Semantically Aligned Discrete Representations}

\author{
Souptik Sen\textsuperscript{1}\quad
Raneen Younis\textsuperscript{1,2}\quad
Zahra Ahmadi\textsuperscript{1,2}\\[0.5em]
\textsuperscript{1}Peter L. Reichertz Institute for Medical Informatics,
Hannover Medical School, Germany\\
\textsuperscript{2}Lower Saxony Center for AI and Causal Methods in Medicine (CAIMed),
Hannover, Germany\\
{\tt\small \{Sen.Souptik, Younis.Raneen, Ahmadi.Zahra\}@mh-hannover.de}
}

\begin{document}
\maketitle
\begin{abstract}
Multimodal learning seeks to integrate information across diverse sensory sources, yet current approaches struggle to balance cross-modal generalizability with modality-specific structure. Continuous (implicit) methods preserve fine-grained priors but render generalization challenging, while discrete (explicit) approaches enforce shared prototypes at the expense of modality specificity. 
We introduce \textbf{CoDAAR}\footnote{Our code is available at \url{https://github.com/EMuLeMultimodal/CoDAAR}} (\textbf{C}ross-m\textbf{o}dal \textbf{D}iscrete \textbf{A}lignment \textbf{A}nd \textbf{R}econstruction), a novel framework that resolves this long-standing trade-off by establishing semantic consensus across modality-specific codebooks through index-level alignment. This design uniquely allows CoDAAR to preserve modality-unique structures while achieving generalizable cross-modal representations within a unified discrete space.
CoDAAR combines two complementary mechanisms: Discrete Temporal Alignment (DTA), which enables fine-grained temporal quantization, and Cascading Semantic Alignment (CSA), which promotes progressive cross-modal semantic agreement. Together, they establish a competition-free unified representation space. Trained with self-supervised reconstruction objectives on paired multimodal sequences, CoDAAR demonstrates robust cross-modal and cross-domain generalization.
Across Cross-Modal Generalization benchmarks, including event classification, localization, video segmentation, and cross-dataset transfer, CoDAAR achieves state-of-the-art performance, establishing a new paradigm for discrete and generalizable multimodal representation learning. 
\end{abstract}
\vspace{-20pt}

\section{Introduction}
\label{sec:intro}
Humans perceive the world through a continuous stream of multimodal sensory inputs: auditory, visual, and linguistic. Yet, our cognitive system encodes these experiences as discrete conceptual units rather than continuous representations. For example, a video of a departing train is abstracted into language tokens like ``train'', ``moving'', ``station'', allowing our mind to decompose multimodal stimuli into symbolic components. This discretized architecture enables cross-modal knowledge transfer; for instance, understanding ``train" in text supports immediate recognition in visual or auditory contexts, supporting higher-order reasoning and generalization across modalities.

\begin{figure}[t]
\resizebox{0.85\columnwidth}{!}{%
  \centering
  % \begin{tikzpicture}
    % \node[draw, dotted, thick, line width=1pt, inner sep=5pt, rounded corners=5pt] {
      \includegraphics[width=\linewidth,keepaspectratio]{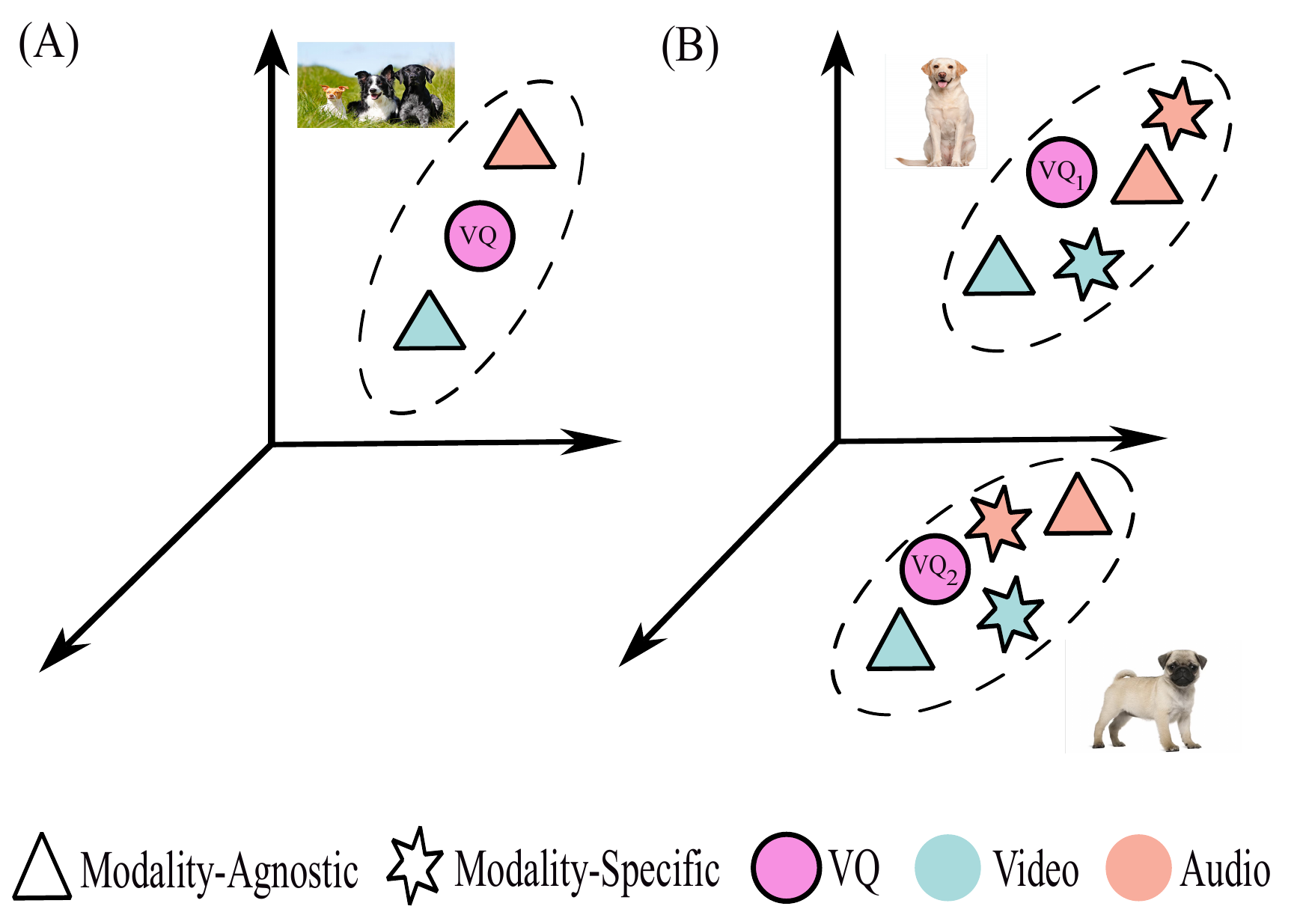}
      }
    % };
  % \end{tikzpicture}
  \caption{(A) Current SOTA unified multimodal representations map only modality-agnostic features to discrete codes. (B) Our proposed representation maps both (i) modality-agnostic and (ii) modality-specific features within a unified space.}
  \label{fig:intro}
  \vspace{-13pt}
\end{figure}
Inspired by this cognitive principle, recent studies have explored unified discrete representations for cross-modal generalization and reasoning. Discrete spaces compress high-dimensional multimodal features into finite sets of generalizable latent codes~\cite{oord2018neural, liu2021cross}, enabling tasks such as visual question answering~\cite{Antol2015VQA, Lei2018TVQA, Li2022AVQA}, query-based segmentation~\cite{Gavrilyuk2018ActorActionSentence, Seo2020URVOS}, and audio-visual event localization~\cite{AVE, Xu2020CMRAN}. Existing methods generally follow two paradigms: (i) \textit{Implicit} continuous models~\cite{DBLP:journals/corr/abs-1807-03748, Radford2021CLIP, CPSP, Akbari2021VATT}, which project modalities into distinct continuous embedding spaces and align them contrastively; and (ii) \textit{Explicit} discrete approaches~\cite{CODIS, TURN, liu2021cross, xia2023achieving, HuangWang25}, which quantize multimodal features via shared codebooks or prototypes.
Implicit models offer flexibility and preserve modality uniqueness but lack generalizability, while explicit methods often lose modality-specific structure by enforcing a shared generalizable space. Furthermore, most discrete frameworks~\cite{CODIS, TURN} are coarse-grained, collapsing sequential features into single prototypes before quantization, discarding temporal semantics. As a result, they remain limited to simple retrieval tasks~\cite{CODIS, TURN} or struggle to retain modality-specific cues needed for complex downstream reasoning~\cite{xia2023achieving, huang2025opensetcrossmodalgeneralization, FCID}.

The significance of preserving modality-specific semantics becomes clear in unconstrained real-world scenarios. Distinguishing visually similar entities, such as a pug versus a labrador, requires nuanced visual cues like fur texture, ear shape, or gait, beyond the generic concept of ''dog''~\cite{Li2022AVQA, Seo2020URVOS} (\cref{fig:intro}). Similarly, in audio-visual data, differentiating a violin from a viola depends on the subtle acoustic timbre and visual morphology of these instruments that coarse semantic alignment alone cannot capture~\cite{AVE, Xu2020CMRAN}. Fine-grained discrete codebook methods~\cite{liu2021cross,xia2023achieving, HuangWang25, huang2025opensetcrossmodalgeneralization, FCID} that rely on a unified quantization encounter a fundamental challenge: \textit{representation competition}. High-variance modalities, primarily vision with its rich spatio-temporal complexity, dominate the shared codebook, biasing codeword positioning towards visual semantics, simultaneously relegating low-variance modalities (e.g., audio) to suboptimal regions~\cite{Liang2022MindTheGap}. This disparity results in an unfavourable trade-off: sacrificing modality-specific structural priors for cross-modal generalization, or preserving modal heterogeneity at the cost of generalizability ~\cite{Li2022AVQA, AVS, Gavrilyuk2018ActorActionSentence}. This raises a central question: \emph{How can we construct multimodal representations that preserve comprehensive modality-specific information while achieving generalizable cross-modal representations?}

To address these challenges, we propose the \textbf{C}ross-M\textbf{o}dal \textbf{D}iscrete \textbf{A}lignment \textbf{A}nd \textbf{R}econstruction (\textbf{CoDAAR}) architecture. CoDAAR introduces modality-specific codebooks instead of a single shared discrete space, directly reducing representational competition. Each modality preserves its intrinsic structural priors within its respective codebook, whereas semantic alignment and generalization arise through index-level correspondence across codebooks. Our architecture comprises two key components: (1) \textbf{\textit{Discrete Temporal Alignment} (DTA)} module handles fine-grained temporal semantics via cross-modal exponential moving average updates at synchronized timeframes across modalities. Each modality codeword aggregates weighted contributions from both fine-grained self-modal features (dominant) and time-synchronized cross-modal streams (auxiliary), thereby balancing the influence of high-variance modalities. (2) \textbf{\textit{Cascading Semantic Alignment} (CSA)} module enforces semantic consensus at the codebook index level by hierarchically aligning corresponding codewords across modalities. This module gravitates the modal-specific codewords at the same index towards a common semantic mean, ensuring that identical indices across codebooks represent similar semantics. This design combines the strengths of both implicit and explicit approaches, preserving modality-specific cues while achieving generalizable cross-modal representations. Our main contributions are as follows:
\begin{itemize}
    \item We introduce \textbf{CoDAAR}, a discrete multimodal alignment framework with two novel modules: \textbf{DTA} for fine-grained temporal quantization through synchronized cross-modal updates, and \textbf{CSA} for hierarchical index-level semantic alignment across modality-specific codebooks.
    \item Our extensive experiments demonstrate cross-modal generalization under scarce annotation and variable labeling costs across modalities, on the Cross-Modal Generalization (CMG) benchmarks~\cite{xia2023achieving}. This evaluation assesses CoDAAR's capacity for zero-shot transfer between labeled and unlabeled modalities. CoDAAR achieves state-of-the-art performance across diverse downstream tasks involving unseen modalities.
\end{itemize}

\begin{figure*}[t]
  \centering
  \begin{subfigure}{0.50\textwidth}
    \centering
    % \tikz[baseline=(current bounding box.center)]{
      % \node[draw, dotted, thick, rounded corners=5pt,  inner sep=4pt] {
        \includegraphics[width=0.95\linewidth,height=0.3\textheight,keepaspectratio]{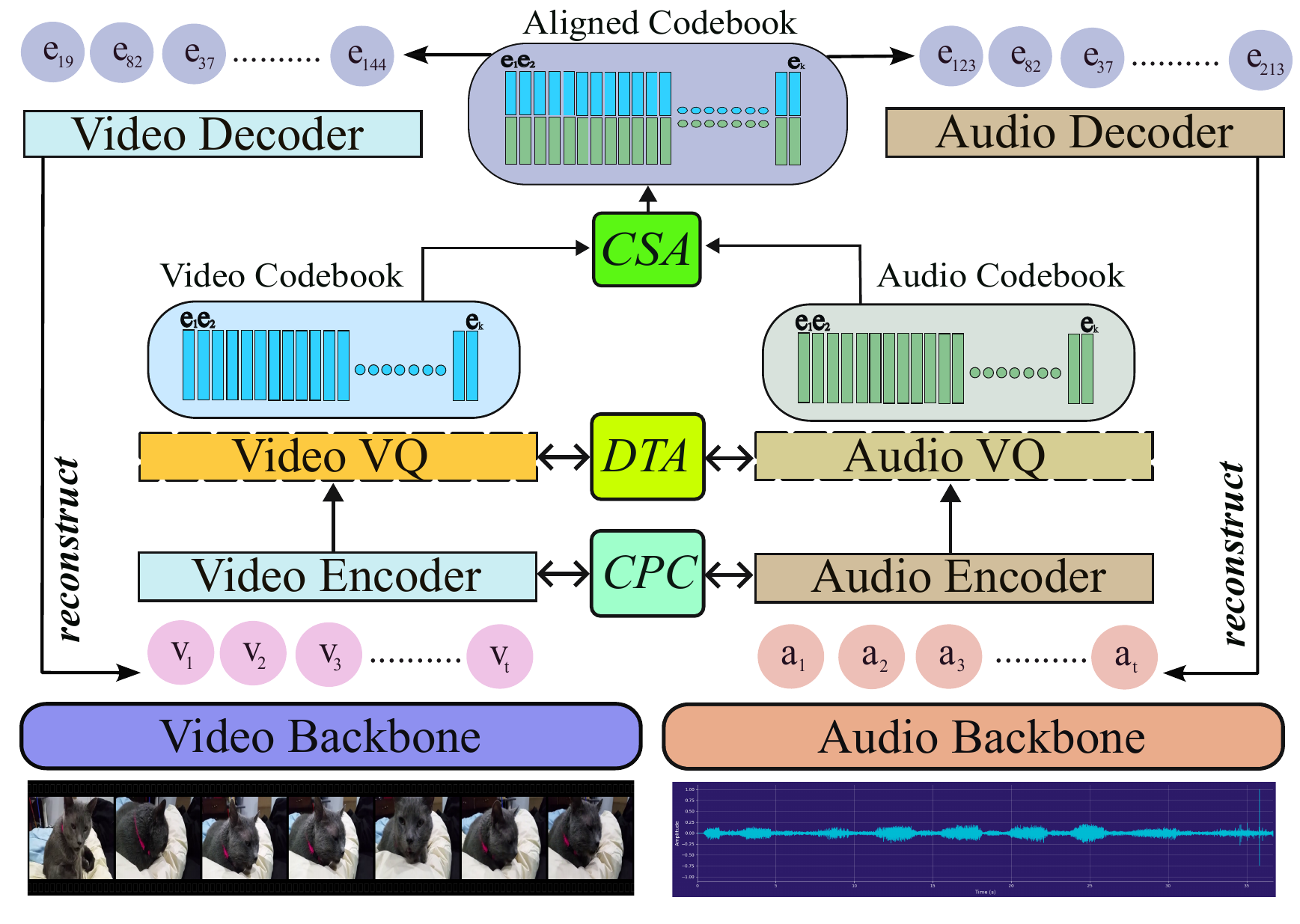}
      % };
    % }
    \caption{Model architecture.}
    \label{fig:sub:model-arch}
  \end{subfigure}
  \hfill
  \begin{subfigure}{0.49\textwidth}
    \centering
    % \tikz[baseline=(current bounding box.center)]{
      % \node[draw, dotted, thick, rounded corners=5pt,  inner sep=4pt] {
        \includegraphics[width=0.95\linewidth,height=0.4\textheight,keepaspectratio]{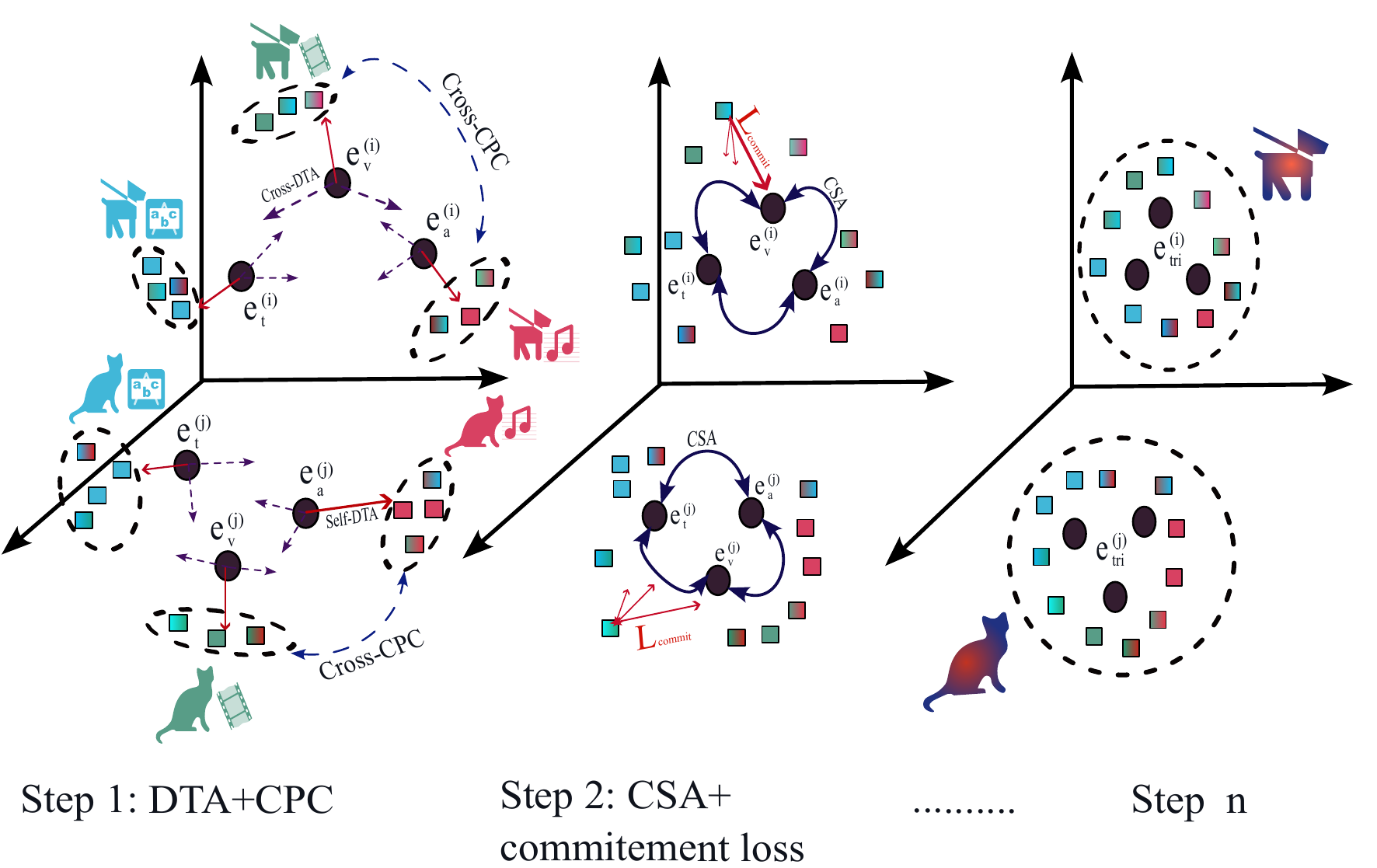} 
      % };
    % }
    \caption{Cross-modal Discrete Alignment}
    \label{fig:sub:seq_ema}
  \end{subfigure}
  \caption{The overview of our proposed CoDAAR framework. (a) Model architecture. (b) The Cross-modal Discrete Alignment mechanism, comprising Discrete Temporal Alignment (DTA), CPC loss, Cascading Semantic Alignment (CSA), and commitment loss.}
  \label{fig:model-overview}
\end{figure*}
% \vspace{-5pt}
\section{Related Work}
\label{sec:Related Works}
\subsection*{Implicit Multimodal Representations}
Implicit multimodal methods align modalities within continuous embedding spaces via contrastive learning~\cite{DBLP:journals/corr/abs-1807-03748, Radford2021CLIP}, spanning vision-language~\cite{Radford2021CLIP, You2022MSCLIP, CLIP4CLIP, UNITER, VLmixer}, audio-visual~\cite{CPSP, unified_space_1, ManifoldLearning, jenni2023audiovisualcontrastivelearningtemporal}, video-audio-text~\cite{Akbari2021VATT, Zorro, wang2024extending}, and speech-text~\cite{SpeechT5, UniSpeech, Mingxuan2021} combinations through contrastive objectives, modality-agnostic transformers, and cross-modal knowledge distillation~\cite{Andonian2022RCD, sarkar2023xkdcrossmodalknowledgedistillation, CTCarch}. While achieving strong cross-modal semantic consistency, their unbounded continuous embeddings inherently restrict generalizability and lack explicit structural similarity across modalities, inhibiting cross-modal generalization.
% \vspace{-5pt}
\subsection*{Explicit Multimodal Representations}
Explicit methods achieve generalization through discrete quantization using shared codebooks~\cite{oord2018neural, liu2021cross} or prototypes~\cite{CODIS} to enable explicit cross-modal similarity. Coarse-grained approaches~\cite{CODIS, TURN} use Optimal Transport or self-cross-reconstruction for prototype alignment but lose temporal semantics by condensing sequences into single vectors, restricting them to basic retrieval tasks. Recent fine-grained methods~\cite{liu2021cross, xia2023achieving, FCID, HuangWang25, huang2025opensetcrossmodalgeneralization} preserve temporal structure through frame-level quantization, improving performance on complex applications~\cite{Li2022AVQA, AVS, huang2025opensetcrossmodalgeneralization}. However, their unified codebook designs create fundamental limitations: high-variance modalities dominate codeword optimization while modality-specific structural priors diminish~\cite{xia2023achieving, huang2025opensetcrossmodalgeneralization, FCID}. Our approach addresses these constraints using modality-specific codebooks that preserve structural priors while achieving semantic consistency through index-level alignment across codebooks.
% trying to get the best of both implicit and explicit algorithms.
% \vspace{-7.5pt}
\section{Cross-Modal Generalization Task Definition}\label{sec:cmg}

The \textit{Cross-Modal Generalization} (CMG) task was introduced in \cite{xia2023achieving} to assess whether multimodal models can learn modality-invariant discrete representations. This task measures how supervision obtained from one modality transfers to another during downstream evaluation. Let $m_1, m_2 \in \{a, v, t\}$ denote audio, video, and text modalities with $m_1 \neq m_2$. During downstream training, the model learns representations for input samples $\mathbf{x}_i^{m_1}$ and their respective labels $\mathbf{y}_i^{m_1}$, using a modality-specific encoder $\Psi^{m_1}(\cdot)$ and a modality-invariant task-head decoder $\mathcal{G}(\cdot)$. The encoder produces continuous embeddings, which are then mapped to discrete latent codes using a vector-quantization operator $VQ(\cdot)$. The decoder is trained on top of these codes using an evaluation loss $\mathcal{E_L}(\cdot)$. At test time, the decoder is evaluated on samples $\mathbf{x}_i^{m_2}$ and labels $\mathbf{y}_i^{m_2}$ from a different modality $m_2$. The parameters of both $\Psi^{m_1}$ and $\Psi^{m_2}$ stay static throughout training and testing, while only the parameters of $\mathcal{G}(\cdot)$ are updated during training. Performance reflects the extent to which the learned discrete space aligns heterogeneous modalities and supports zero-shot knowledge transfer from $m_1$ to $m_2$. A complete notation library is provided in the supplementary material.

% \vspace{-1pt}
\section{CoDAAR Architecture}
\subsection{Cross-Modal Discrete Alignment Framework}

We introduce \textbf{C}ross-M\textbf{o}dal \textbf{D}iscrete \textbf{A}lignment \textbf{A}nd \textbf{R}econstruction (CoDAAR), a framework that constructs a unified discrete latent space for fine-grained cross-modal and cross-domain generalization. Given a paired multimodal dataset $\mathbf{X} = \{(x_i^a, x_i^v, x_i^t)\}_{i=1}^{N}$ with aligned audio, video, and text sequences, CoDAAR assigns each modality $m$ a codebook $\mathbf{E}_m \in \mathbb{R}^{K \times D}$ containing $K$ codewords. Each modality's input is encoded into continuous semantic embeddings, then discretized via these codebook indices. The key innovation is aligning these indices such that identical indices across modalities represent the same latent concept, yielding a unified discrete vocabulary $\mathbf{E} = [\mathbf{E}_a ; \mathbf{E}_v ; \mathbf{E}_t]$. This aligned index space enables cross-modal knowledge transfer while preserving modality-specific structural cues in respective codebooks through two complementary objectives: reconstruction for capturing modality-specific details and alignment for enforcing cross-modal semantic coherence, supporting evaluation under the CMG protocol (Section~\ref{sec:cmg}). Domain generalization emerges through unsupervised pre-training on these structural and semantic patterns. Figure~\ref{fig:sub:model-arch} illustrates the architecture. 
% Notation library can be found in the appendix.

% \vspace{-0.5pt}
\subsubsection{VQ-VAE Backbone}\label{subsubsec:baseline}
Given paired multimodal sequences $\mathbf{X}=\{(x_i^a, x_i^v, x_i^t)\}_{i=1}^{N}$, the core backbone module follows a vector-quantized reconstruction pipeline. For each modality $m \in \{a,v,t\}$, an encoder $\Psi^{m}(\cdot)$ maps inputs $x_i^m$ to a continuous latent representation $z_i^m = \Psi^{m}(x_i^m) \in \mathbb{R}^{T \times D}$, where $T$ and $D$ denote the temporal length and feature dimension. Each modality maintains a codebook $\mathbf{E}_m \in \mathbb{R}^{K \times D}$ with $K$ codewords of the same dimensionality as the latent embeddings. Encoder embeddings are discretized to these codebooks via a nearest-neighbour lookup. Each vector frame $z_{i,t}^m,t \in \{1{:}T\}$, is mapped to a quantized embedding, $\hat{z}_{i,t}^m = \mathrm{VQ}(z_{i,t}^m) = \mathbf{e}_m(k)$, where the index $k$ is selected by$\quad k = \arg\min_{j}\|z^m_{i,t} - \mathbf{e}_m(j)\|_2^2$. For reconstruction, the quantized embeddings $\hat{z}_i^m$ are concatenated with a modality-specific projection $P_m(x_i^m)$ of the input features, and passed through a decoder $\mathcal{D}_m(\cdot)$ to obtain $\tilde{x}_i^{m}=\mathcal{D}_m([\hat{z}_i^{m};P_m(x_i^{m})])$.
Training follows the standard VQ-VAE objective~\cite{oord2018neural}, combining reconstruction, quantization, and commitment terms:
\begin{equation}
\mathcal{L} =
\underbrace{\|x^m_i - \tilde{x}^m_i\|_2^2}_{\text{reconstruction loss}}
+ 
\underbrace{\|\mathrm{sg}[z^m_{i}] - \mathbf{e}_m\|_2^2}_{\text{VQ loss}}
+ 
\underbrace{\beta \|z^m_{i} - \mathrm{sg}[\mathbf{e}_m]\|_2^2}_{\text{commitment loss}},
\label{eq:loss_full}
\end{equation}
where $\mathrm{sg}[\cdot]$ is the stop-gradient operator and $\beta = 0.25$ in all experiments~\cite{oord2018neural}. Following ~\cite{TURN, oord2018neural}, we construct the codebook via Exponential Moving Average (EMA) operation instead of the explicit VQ loss, improving stability and preventing codebook collapse. Without additional constraints, this core formulation learns modality-specific codebooks that remain semantically misaligned across modalities. We use this backbone to isolate the effects of our alignment mechanism, ensuring that improvements arise solely from cross-modal correspondence.

\subsubsection{CPC-based Cross-Modal Information Maximization}
To inject multimodal semantics into our unimodal encoders $\Psi^{m}(\cdot)$, we adopt a Contrastive Predictive Coding (CPC) objective~\cite{DBLP:journals/corr/abs-1807-03748, xia2023achieving} that makes each encoder cross-modally aware. For a modality pair $(a,b)$, a unidirectional LSTM summarizes past embeddings of modality $a$ into a context vector $c_t^{a}$, which is optimized to predict future embeddings of modality $b$. This incorporates fine-grained temporal cross-modal cues directly into the unconstrained unimodal streams, improving modal transferability during downstream inference. We apply the objective symmetrically across ordered modality pairs and average their contributions:
\begin{equation}
\mathcal{L}_{\text{CPC}}
= -\frac{1}{H}\sum_{h=1}^{H}\log
\frac{\exp\!\big((z^{b}_{t+h})^{\top} W_{h}^{a} \, c^{a}_{t}\big)}
{\sum_{j}\exp\!\big((z^{b}_{j})^{\top} W_{h}^{a} \, c^{a}_{t}\big)},
\label{eq:cpc_loss}
\end{equation}
where $c^{a}_{t}$ is the modality-$a$ context at time $t$; $z^{b}_{t+h}$ is the future latent of modality $b$; $W_h^{a}$ is a step-specific linear projection; the denominator aggregates one positive and sampled negatives $\{z^{b}_{j}\}$ from modality $b$.

\subsubsection{Cross-Modal Discrete Alignment}
With our CPC-induced cross-modally aware encoders $\Psi^{m}(\cdot)$ in place, we introduce a three-fold alignment to learn our codebooks. \textit{DQA} aligns codebook index selection across modalities; \textit{DTA} temporally aligns the construction of these selected codewords via cross-modal EMA; and \textit{CSA} applies a cascading shift to these constructed codewords to enforce cross-index semantic alignment.
\subsubsection*{Discrete Quantization Alignment (DQA)}

To ensure index utilization of an embedding \(i\) is consistent across modalities \(m\!\in\!\{a,v,t\}\), we employ the Cross-Modal Code Matching (CMCM)~\cite{liu2021cross} loss, extending it to our modality-specific codebooks \(\{\mathbf{E}_m\}_{m\in\{a,v,t\}}\). For each paired embedding $(z_i^a, z_i^v, z_i^t)$, we compute the sequence-level code-usage distributions $p_m^{\,i}$ over indices $k \in \{1,\dots,K\}$ and align these distributions across modalities. We match the embedding distributions of corresponding indices by contrasting distributions from modality-paired sample embeddings while treating non-matching samples in the batch as negatives. This encourages the model to assign the same index to the same embedding instance across modalities, leading to consistent nearest-neighbour lookups even though each modality quantizes through a different codebook. DQA therefore promotes cross-modal index usage synchronization. However, it does not guarantee semantically aligned codebook constructions. The DQA loss equation can be found in the supplementary material.
\vspace{-1pt}
% The negative pairs improve index discrimination and reduce accidental overlap between unrelated samples.
% \begin{equation}
% \mathcal{L}_{\text{CMCM}}
% = -\frac{1}{N}\sum_{i}\!
% \log\!
% \frac{\exp\!\big(\langle p_a^{\,i}, \log p_b^{\,i}\rangle + \langle p_b^{\,i}, \log p_a^{\,i}\rangle\big)}
% {\sum_{j}\exp\!\big(\langle p_a^{\,i}, \log p_b^{\,j}\rangle + \langle p_b^{\,j}, \log p_a^{\,i}\rangle\big)},
% \label{eq:cmcm}
% % \end{equation}
% where \(p_m^{\,i}\!\in\!\Delta^{K-1}\) is the time-averaged soft assignment over indices \(k=1..K\) for modality \(m\) and sample \(i\).
\subsubsection*{Discrete Temporal Alignment (DTA)}
We introduce DTA to enable fine-grained temporal quantization and to \emph{align} how selected codewords are \emph{constructed} over training iterations. We treat codewords as k-means centroids for the latent encoder embeddings and update them using a temporally aligned, cross-modal EMA. For each timestep of a paired sample embedding, the centroid of modality $m$ moves toward a convex combination of its own quantized embeddings and the paired embeddings from the other modalities at that same timestep. Cross-modally aware encoders produce temporally aligned embeddings with comparable semantics at identical timesteps within a sample. This allows our cross-modal EMA updates to aggregate the underlying aligned semantics.

\noindent\textbf{Indexing:}  
A mini-batch contains $B$ samples indexed by $i \in \{1{:}B\}$, each with $T$ timesteps.  
For modality $m \in \{a,v,t\}$, the latent encoder embedding is $\mathbf{z}_{i}^m \in \mathbb{R}^{T \times D}$.  
% Alignment is performed \emph{per sample and timestep} across the triplet $(\mathbf{x}_{a,i,t}, \mathbf{x}_{v,i,t}, \mathbf{x}_{t,i,t})$ for the same $(i,t)$.

\noindent\textbf{Hard assignment (per codeword $k$):}  
For modality $m$ with codebook $\mathbf{E}_m=\{\mathbf{e}_m(k)\}_{k=1}^{K}$, each embedding frame $ \mathbf{z}_{i,t}^m, t \in \{1{:}T\}$, selects exactly one codeword via nearest-neighbour lookup:
\begin{equation}
\label{eq:hard_assign}
\begin{aligned}
\pi_m(i,t) &= \operatorname*{arg\,min}_{j\in\{1{:}K\}}
\left\lVert \mathbf{z}_{i,t}^m-\mathbf{e}_m(j)\right\rVert_2^2, \\
q_{m,i,t}(k) &= \mathbb{1}\!\left[k=\pi_m(i,t)\right].
\end{aligned}
\end{equation}
Stacking \(q_{m,i,t}(k)\) over \((i,t)\) produces a binary responsibility tensor with one-hot rows per \((i,t)\).

\noindent\textbf{Temporally aligned accumulators:}
For \(\{n,p\}=\{a,v,t\}\setminus\{m\}\), we define EMA inputs that aggregate matched \((i,t)\) assignments:
\begin{equation}
\label{eq:temporal_ema_defs_hard}
\begin{aligned}
\mathbf{U}_m^{\text{self}}(k)  &\;=\; \sum_{i=1}^{B}\sum_{t=1}^{T} q_{m,i,t}(k)\,\mathbf{z}_{i,t}^m, \\
\mathbf{U}_m^{\text{cross}}(k) &\;=\; \sum_{i=1}^{B}\sum_{t=1}^{T} q_{m,i,t}(k)\,\big(\mathbf{z}_{i,t}^n+\mathbf{z}_{i,t}^p\big).
\end{aligned}
\end{equation}

\noindent\textbf{EMA update (per codeword \(k\)):}
We use $\lambda_{\text{self}}{=}0.6$, $\lambda_{\text{cross}}{=}0.2$ (two cross terms; $0.6{+}0.2{+}0.2=1$), decay $\rho\!\in\!(0,1)$, and $\varepsilon\!>\!0$. The update is:
\begin{equation}
\label{eq:temporal_ema_update}
\begin{aligned}
\mathbf{H}_m(k)                &\;=\; \lambda_{\text{self}}\mathbf{U}_m^{\text{self}}(k)
                                  + \lambda_{\text{cross}}\mathbf{U}_m^{\text{cross}}(k),\\
\mathbf{S}_m^{\text{new}}(k)   &\;=\; \rho\,\mathbf{S}_m(k) + (1-\rho)\,\mathbf{H}_m(k),\\
C_m^{\text{new}}(k)            &\;=\; \rho\,C_m(k) + (1-\rho)\!\sum_{i=1}^{B}\sum_{t=1}^{T} q_{m,i,t}(k),\\
\mathbf{e}_m^{\text{new}}(k)   &\;=\; \dfrac{\mathbf{S}_m^{\text{new}}(k)}{\,C_m^{\text{new}}(k)+\varepsilon\,}.
\end{aligned}
\end{equation}
Here, $\mathbf{S}_m(k)$ tracks the EMA-weighted sum of assigned features (self and time-aligned cross-modal), $C_m(k)$ tracks the EMA-weighted assignment count, and $\mathbf{e}_m^{\text{new}}(k)$ is the resulting centroid. The weights $(0.6,0.2,0.2)$ form a convex partition. The self-term anchors each codeword in its own modality. The cross-terms incorporate time-aligned evidence from the paired modalities at the same $(i,t)$, acting as soft teachers. Updating the codebooks separately for each modality with dominant self-anchors, unlike methods with a shared codebook~\cite{xia2023achieving,huang2025opensetcrossmodalgeneralization}, prevents competition in a single shared space and avoids dominance by any one modality. 
\vspace{-1pt}
\subsubsection*{Cascading Semantic Alignment (CSA)}
\begin{figure}[t]
  \centering
  % \tikz[baseline=(current bounding box.center)]{
    % \node[draw, dotted, thick, rounded corners=5pt, inner sep=3pt] {
      \includegraphics[width=0.95\linewidth]{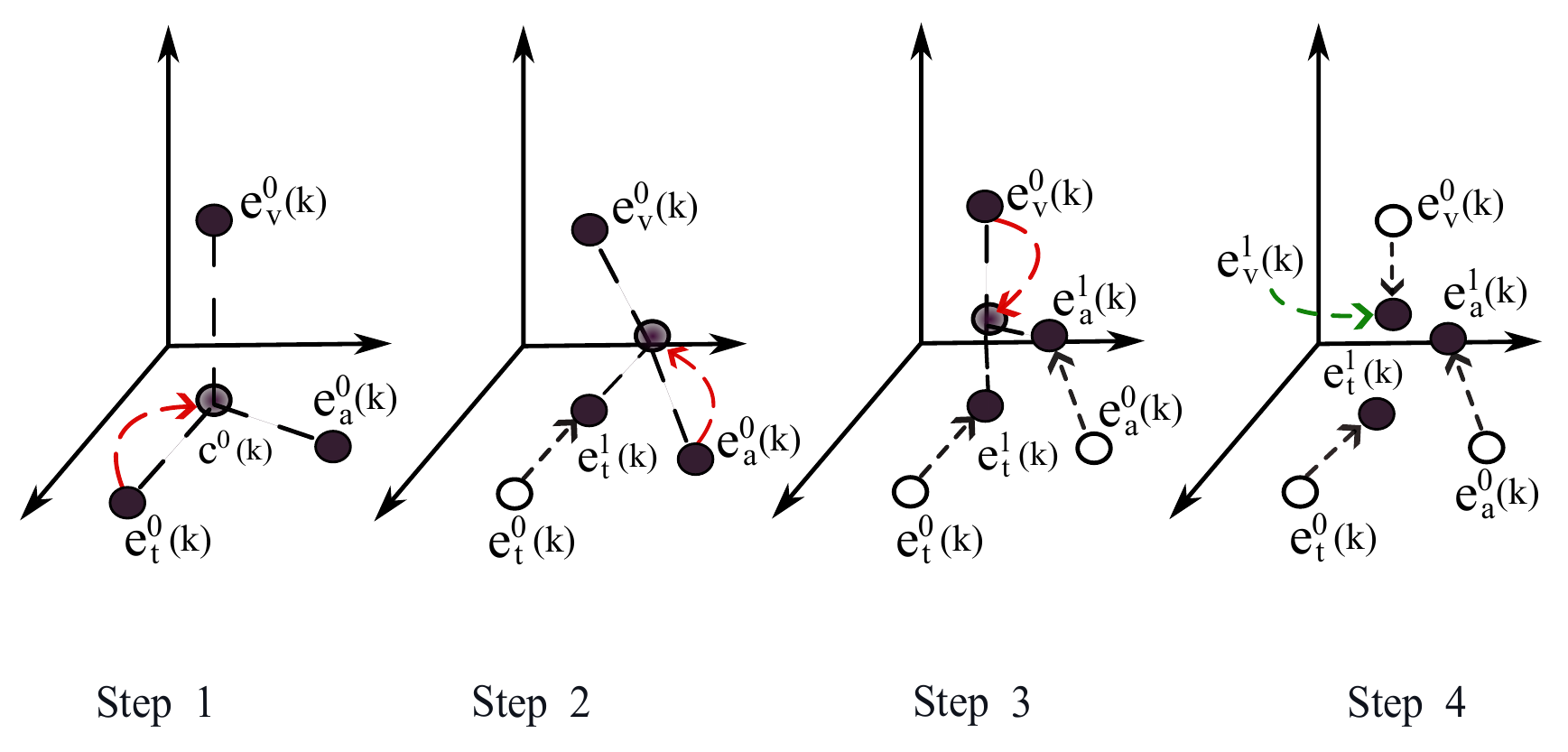}
    % };
  % }
  \caption{Cascading Semantic Alignment Visualized}
  \label{fig:cascade}
\end{figure}

Although the unimodal codebooks already mix modality-specific and shared cues through DTA, index-level semantic misalignment arises from independent initialization across $\mathbf{E}_a$, $\mathbf{E}_v$, and $\mathbf{E}_t$ and modality biases. To enforce semantic consistency across modality-specific codebooks, we introduce this hierarchical cascade module that gravitates the three centroids at index $k$ toward a shared multimodal semantic mean anchored by $\mathbf{c}^0(k)$ (Eq.~\ref{eq:cm_center}). The cascade mitigates the misalignment by applying a sequential T$\rightarrow$A$\rightarrow$V centroid shift (Fig.~\ref{fig:cascade}), ordered by increasing relative semantic capacity: text (global semantics) $<$ audio (temporal semantics) $<$ video (spatio-temporal semantics). Text is first pulled toward the cross-modal anchor $\mathbf{c}^0(k)$; audio then bridges text and video; video finalizes the consensus while retaining the most detail (Fig.~\ref{fig:cascade}). Intuitively, the multimodal semantic mean tends to lie closer to the higher-capacity stream, i.e., video, which contributes additional spatial detail.
Let $\mathbf{e}_v^0(k),\mathbf{e}_a^0(k),\mathbf{e}_t^0(k)\!\in\!\mathbb{R}^{D}$ be post-DTA centroids at index $k$ in a mini-batch iteration. We define $\mathbf{c}^0(k)$, the geometric centroid of the 3 codewords, as the cross-modal semantic anchor for index $k$.
\begin{equation}
\label{eq:cm_center}
\mathbf{c}^0(k)=\tfrac{1}{3}\big(\mathbf{e}_v^0(k)+\mathbf{e}_a^0(k)+\mathbf{e}_t^0(k)\big).
\end{equation}
We apply cascading equal-weight updates to the modality centroids:
\begin{equation}
\label{eq:cascade_short}
\begin{aligned}
\mathbf{e}_t^1(k) &= \mathbf{c}^0(k), \\
\mathbf{e}_a^1(k) &= \tfrac{1}{3}\!\left(\mathbf{e}_a^0(k)+\mathbf{e}_v^0(k)+\mathbf{e}_t^1(k)\right), \\
\mathbf{e}_v^1(k) &= \tfrac{1}{3}\!\left(\mathbf{e}_v^0(k)+\mathbf{e}_a^1(k)+\mathbf{e}_t^1(k)\right).
\end{aligned}
\end{equation}
Thus, the modality-specific centroids at index $k$ shift toward distinct multimodal locations in the representation space rather than collapsing into a single trimodal point. This allows the centroids to share a unified semantic meaning while retaining modality-specific characteristics, with video (last) retaining the most. Learned mixing weights destabilize our EMA update and allow one modality to dominate, causing codebook collapses. Therefore, we use fixed, non-negative update weights in Eq.~\ref{eq:cascade_short}; that sum to $1$ and create progressive centroids. This keeps each update inside the convex hull of its contributors and stabilizes codebook construction. As training proceeds, and \textbf{DTA} aggregates more paired embeddings, the three codewords at index $k$ move toward semantic consensus in an independent yet coordinated manner through this complementary cascade module. The commitment loss (Eq.~\ref{eq:commitment}) then pulls multimodal latents toward their assigned centroids, forming \emph{distinct semantic spheres} in the multimodal representation space (Fig.~\ref{fig:sub:seq_ema}).

\subsubsection{Pretraining Objectives}

\noindent\textbf{Cross-modal Commitment loss:}
For modality $m\!\in\!\{a,v,t\}$, let $z_{i}^{m}=\Psi^{m}(x_{i}^{m})$ be the encoder embedding of sample $i$, $z_{i,t}^{m}$, $t \in \{1{:}T\}$, the embedding frame, and let $k_{i,t}$ be the index selected by the quantizer (nearest code) in modality $m$, with codewords $\mathbf{e}_m(k)\in\mathbf{E}_{m}$.
Using the stop–gradient operator $\mathrm{sg}[\cdot]$, we define
% \begin{equation}
% \label{eq:commitment}
% \mathcal{L}_{\text{commit}}^{\,m}
% =\sum_{t=1}^{T}\!\Big[\beta\|z_{i,t}^{m}-\mathrm{sg}(e_{k_{i,t}}^{m})\|_2^{2}
% +\tfrac{\beta}{2}\!\sum_{n\neq m}\!\|z_{i,t}^{m}-\mathrm{sg}(e_{k_{i,t}}^{n})\|_2^{2}\Big].
% \end{equation}
\begin{equation}
\label{eq:commitment}
\begin{aligned}
\mathcal{L}_{\text{commit}}^{\,m} = \sum_{t=1}^{T}\Big[&\beta\|z_{i,t}^{m}-\mathrm{sg}[e_m(k_{i,t})]\|_2^{2}\\
&+\tfrac{\beta}{2}\!\sum_{n\neq m}\!\|z_{i,t}^{m}-\mathrm{sg}[e_n(k_{i,t})]\|_2^{2}\Big].
\end{aligned}
\end{equation}
where $n\!\in\!\{a,v,t\}$, and $\beta=0.25$.
The first term makes each encoder embedding \emph{commit} to its own selected codeword; the (weaker) cross terms ($\beta/2$) softly pull it toward the same-index centroids of the other modalities, encouraging index-level alignment.

\noindent\textbf{Cross-modal reconstruction:}
After aligning the cross-modal codeword indices, we get $\hat{\mathbf{z}}^{m}_i\in\mathbb{R}^{T\times D}$ as the quantized latent of modality $m\in\{a,v,t\}$. Thus, we can define the concatenated tri-modal code as: $\hat{\mathbf{z}}^{\mathrm{tri}}_i = [\hat{\mathbf{z}}^{a}_i; \hat{\mathbf{z}}^{v}_i; \hat{\mathbf{z}}^{t}_i] \in \mathbb{R}^{T\times 3D}$.
Each modality is reconstructed from the \emph{same} trimodal code, instead of the unimodal codes, concatenated with the modality-specific projection as defined in \cref{subsubsec:baseline}:
% \begin{equation}
% \label{eq:recon_cross}
% \begin{aligned}
% \tilde{\mathbf{x}}^{\,m}_i &= \mathcal{D}_m\!\big([\hat{\mathbf{z}}^{\mathrm{tri}}_i \,;\, P_m(\mathbf{x}^{m}_i)]\big),\\[-2pt]
% \mathcal{L}_{\mathrm{rec}}^{\,m} &= \sum_{m\in\{a,v,t\}} \big\lVert \mathbf{x}^{m}_i - \tilde{\mathbf{x}}^{\,m}_i \big\rVert_2^2.
% \end{aligned}
% \end{equation}
\begin{equation}
\label{eq:recon_cross}
\begin{aligned}
\tilde{\mathbf{x}}^{\,m}_i &= \mathcal{D}_m\!\big([\hat{\mathbf{z}}^{\mathrm{tri}}_i \,;\, P_m(\mathbf{x}^{m}_i)]\big),\\[-2pt]
\mathcal{L}_{\mathrm{rec}}^{\,m} &=  \big\lVert \mathbf{x}^{m}_i - \tilde{\mathbf{x}}^{\,m}_i \big\rVert_2^2.
\end{aligned}
\end{equation}
This introduces cross-modal reconstruction gradients that flow to all unimodal encoders $\Psi^{m}(\cdot)$, making them cross-modally aware.

\noindent\textbf{Final Loss:}
Our pre-training minimizes the following composite objective:
\begin{equation}
\label{eq:total_loss}
\begin{aligned}
\mathcal{L}_{\text{total}}
&=\sum_{m\in\{a,v,t\}}\!\big(\mathcal{L}_{\text{rec}}^{\,m}+\mathcal{L}_{\text{commit}}^{\,m}\big)\\
&\quad+\sum_{(m,n)\in\mathcal{P}}\!\big(\mathcal{L}_{\text{CPC}}^{\,m\leftrightarrow n}
+\mathcal{L}_{\text{CMCM}}^{\,m\leftrightarrow n}\big).
\end{aligned}
\end{equation}
Here $\mathcal{P}=\{(a,v),(a,t),(v,t)\}$ denotes modality pairs available in a batch. 
\noindent\textbf{Reset Code:} Following~\cite{xia2023achieving}, we \textit{reset} any inactivated code that is unselected for $N_{\mathrm{re}}$ consecutive batches by reinitializing it from an active code plus small noise, preventing dead codes.
\subsection{Downstream Protocol}
After pretraining, \textit{we freeze all our encoders and codebooks}. Downstream protocols involve training only lightweight task heads with an evaluation loss $\mathcal{E_L}(\cdot)$ following the CMG evaluation setup in Section~\ref{sec:cmg}.
\section{Experiments}

\begin{table*}[t]
  \centering
  \setlength{\tabcolsep}{6pt}
  \renewcommand{\arraystretch}{1.1}
  \resizebox{0.6\textwidth}{!}{%
  \begin{tabular}{l|cc|cc|cc|cc|c}
    \hline
    \multirow{2}{*}{Method}
      & \multicolumn{4}{c|}{\textbf{VGGSound-AVEL 40K}}
      & \multicolumn{4}{c|}{\textbf{VGGSound-AVEL 90K}}
      % & \multirow{2}{*}{Average} \\
      & \multicolumn{1}{c}{Average} \\
    \cline{2-9}
      & \multicolumn{2}{c|}{AVE}
      & \multicolumn{2}{c|}{AVVP}
      & \multicolumn{2}{c|}{AVE}
      & \multicolumn{2}{c|}{AVVP}
      & \\
    \cline{2-9}
      & V$\rightarrow$A & A$\rightarrow$V & V$\rightarrow$A & A$\rightarrow$V
      & V$\rightarrow$A & A$\rightarrow$V & V$\rightarrow$A & A$\rightarrow$V
      & \\
    \hline
    Baseline & 5.1 & 4.5 & 5.5 & 4.3 & 7.5 & 10.2 & 6.2 & 7.6 & 6.4 \\
    CMCM~\cite{liu2021cross} & 32.7 & 36.8 & 41.9 & 45.1 & 30.5 & 33.7 & 38.4 & 43.9 & 37.9 \\
    CODIS~\cite{CODIS} & 20.8 & 26.4 & 35.1 & 37.9 & 29.3 & 31.1 & 33.8 & 36.4 & 31.4 \\
    TURN~\cite{TURN} & 19.1 & 24.3 & 36.9 & 39.3 & 28.5 & 32.2 & 32.5 & 37.6 & 31.3 \\
    DCID~\cite{xia2023achieving} & \textbf{47.7} & \textbf{52.3} & \textbf{64.0} & 65.6 & 34.8 & 49.0 & 59.7 & 64.8 & 54.7 \\
    MICU~\cite{huang2025opensetcrossmodalgeneralization} & 47.2 & 51.4 & 38.4 & 33.5 & 33.0 & 37.7 & 42.9 & 46.1 & 41.3 \\
    \textbf{CoDAAR (ours)} & 47.6 & 49.7 & 63.6 & \textbf{72.4} & \textbf{43.3} & \textbf{49.4} & \textbf{60.0} & \textbf{67.2} & \textbf{56.6} \\
    \hline
  \end{tabular}
  }
  \caption{AV setting: comparison with state-of-the-art methods on (i) event classification (AVE, Precision) and (ii) event localization (AVVP, Segment-level Accuracy). V$\rightarrow$A / A$\rightarrow$V denote transfer directions.}
  \label{tab:sota_ave_avvp}
\end{table*}

% ---------- AVT: AVE & AVVP ----------
\begin{table*}[t]
  \centering
  \setlength{\tabcolsep}{6pt}
  \renewcommand{\arraystretch}{1.1}
  \resizebox{0.6\textwidth}{!}{%
  \begin{tabular}{l|cc|cc|cc|cc|c}
    \hline
    \multirow{2}{*}{Method}
      & \multicolumn{4}{c|}{\textbf{VGGSound-AVEL 40K}}
      & \multicolumn{4}{c|}{\textbf{VGGSound-AVEL 90K}}
      % & \multirow{2}{*}{Average} \\
      & \multicolumn{1}{c}{Average} \\
    \cline{2-9}
      & \multicolumn{2}{c|}{AVE}
      & \multicolumn{2}{c|}{AVVP}
      & \multicolumn{2}{c|}{AVE}
      & \multicolumn{2}{c|}{AVVP}
      & \\
    \cline{2-9}
      & V$\rightarrow$A & A$\rightarrow$V & V$\rightarrow$A & A$\rightarrow$V
      & V$\rightarrow$A & A$\rightarrow$V & V$\rightarrow$A & A$\rightarrow$V & \\
    \hline
    Baseline                                   & 6.2 & 5.7 & 7.1 & 6.5 & 8.3 & 9.1 & 8.6 & 9.4 & 7.6 \\
    DCID~\cite{xia2023achieving}               & 54.1  & 55.0  & 63.4  & 71.0  & 43.7  & 50.3  & 64.0  & 64.2  & 58.2  \\
    MICU~\cite{huang2025opensetcrossmodalgeneralization} & \textbf{56.1}  & \textbf{57.1}  & 59.5  & 56.2  & 43.1  & 47.8  & 45.6  & 47.1  & 51.6  \\
    \textbf{CoDAAR (ours)}                      & 52.3 & 55.5 & \textbf{70.8} & \textbf{72.5}
                                               & \textbf{50.8} & \textbf{51.9} & \textbf{69.7} & \textbf{70.4} & \textbf{61.7} \\
    \hline
  \end{tabular}
  }
  \caption{AVT setting:  comparison with state-of-the-art methods on (i) event classification (AVE, Precision) and  (ii) event localization (AVVP, Segment-level Accuracy).}
  \label{tab:sota_avt_ave_avvp}
\end{table*}

% ---------- Zero/Shot Cross-Modal retrieval on MSCOCO and Clotho ----------

\begin{table*}[t]
    \centering
    \setlength{\tabcolsep}{5pt}
    \renewcommand{\arraystretch}{1.1}
    \resizebox{0.78\textwidth}{!}{%
    \begin{tabular}{l|ccc|ccc|ccc|ccc|c}
    \hline
    \multirow{3}{*}{Method}
      & \multicolumn{6}{c|}{\textbf{VGGSound-AVEL 40K}}
      & \multicolumn{6}{c|}{\textbf{VGGSound-AVEL 90K}}
      & \multirow{3}{*}{Average} \\
    \cline{2-13}
      & \multicolumn{3}{c|}{MSCOCO(V$\leftrightarrow$T)}
      & \multicolumn{3}{c|}{Clotho(A$\leftrightarrow$T)}
      & \multicolumn{3}{c|}{MSCOCO(V$\leftrightarrow$T)}
      & \multicolumn{3}{c|}{Clotho(A$\leftrightarrow$T)}
      & \\
    \cline{2-13}
      & R@1 & R@5 & R@10 & R@1 & R@5 & R@10
      & R@1 & R@5 & R@10 & R@1 & R@5 & R@10 & \\
    \hline
    DCID~\cite{xia2023achieving}  & 0.80 & \textbf{5.00} & 8.30 & 2.06 & 9.00 & 16.70 & 0.80 & 4.80 & 8.20 & 2.77 & 11.00 & 20.43 & 7.49 \\
    MICU~\cite{huang2025opensetcrossmodalgeneralization} & \textbf{1.30} & \textbf{5.00} & 8.80 & 2.44 & 10.96 & 18.95 & 0.90 & 4.90 & 9.50 & 3.35 & 10.07 & 17.62 & 7.82 \\
    \textbf{CoDAAR (ours)}                       & 0.80 & 4.40 & \textbf{9.50} & \textbf{3.30} & \textbf{11.57} & \textbf{19.00}
                                                & \textbf{1.30} & \textbf{6.10} & \textbf{10.40} & \textbf{5.64} & \textbf{16.65} & \textbf{24.70} & \textbf{9.45} \\
    \hline
    \end{tabular}
    }
    \captionof{table}{Cross-modal zero-shot retrieval comparison with discrete SOTA methods on MSCOCO (V$\leftrightarrow$T) and Clotho (A$\leftrightarrow$T) datasets.}
    \label{tab:sota_retrieval_main}
\end{table*}

% ---------- AVSBench-S4 video segmentation (A2T / T2A; mIoU & F-score) ----------
\begin{table*}[t]
  \centering
  \setlength{\tabcolsep}{6pt}
  \renewcommand{\arraystretch}{1.1}
  \resizebox{0.6\textwidth}{!}{%
  \begin{tabular}{l|cc|cc|cc|cc|c}
    \hline
    \multirow{2}{*}{Methods}
      & \multicolumn{4}{c|}{\textbf{VGGSound-AVEL 40K}}
      & \multicolumn{4}{c|}{\textbf{VGGSound-AVEL 90K}}
      & \multirow{2}{*}{Average} \\
    \cline{2-9}
      & \multicolumn{2}{c|}{A$\rightarrow$T} & \multicolumn{2}{c|}{T$\rightarrow$A}
      & \multicolumn{2}{c|}{A$\rightarrow$T} & \multicolumn{2}{c|}{T$\rightarrow$A}
      & \\
    \cline{2-9}
      & mIoU & F-score & mIoU & F-score
      & mIoU & F-score & mIoU & F-score
      & \\
    \hline
    DCID~\cite{xia2023achieving}             & 73.3   & 83.3  & \textbf{77.7} & \textbf{86.7}   & 74.8   & 84.7   & \textbf{76.2}   & \textbf{86.3}   & 80.4 \\
    MICU~\cite{huang2025opensetcrossmodalgeneralization} & 73.8   & 84.1   & 76.9   & 86.5   & 76.4   & 86.0   & 75.7   & 85.7   & 80.6 \\
    \textbf{CoDAAR (ours)}                    & \textbf{74.3} & \textbf{84.5} & 75.4 & 86.6
                                             & \textbf{76.5} & \textbf{86.4} & 76.0 & 86.0
                                             & \textbf{80.7} \\
    \hline
  \end{tabular}
  }
  \caption{AVT setting: comparison with state-of-the-art methods on \textbf{AVSBench-S4} query-based video segmentation task, fine-tuned till the 4th downstream epoch. Columns report mIoU and F-score for audio $\!\to\!$ text and text $\!\to\!$ audio.}
  \label{tab:avs_s4}
\end{table*}

\subsection{Evaluation Setup}

% \noindent\textbf{Pretrain:}
We evaluate CoDAAR under two pretraining settings: \textbf{AV} (Audio–Video) and \textbf{AVT} (Audio–Video–Text). Challenging cross-modal and cross-domain transfer experiments use only AVT due to AV's lack of textual semantic grounding. We pretrain on VGGSound-AVEL~\cite{CPSP,vggsound} using 40k and 90k splits, which contain audio, video, and event labels; for AVT, we incorporate text via event-label descriptions from~\cite{xia2023achieving}.

\noindent\textbf{1. Cross-modal Event Classification (AVE)~\cite{AVE}:}  
Each video contains a single event label. We train a global event classifier using one modality (e.g., video) and evaluate zero-shot performance on another (e.g., audio). Precision is reported as the evaluation metric.

\noindent\textbf{2. Cross-modal Event Localization (AVVP)~\cite{AVVP}:}  
This multi-label dataset includes temporally overlapping events. We train a fine-grained event classifier on one modality and test on another, measuring segment-level accuracy to capture temporal alignment and class imbalance.

\noindent\textbf{3. Cross-Modal Zero-Shot Retrieval (MSCOCO~\cite{lin2014microsoft}, Clotho~\cite{drossos2020clotho}):} 
We evaluate zero-shot cross-modal retrieval on MSCOCO for vision$\leftrightarrow$text and Clotho for audio$\leftrightarrow$text by retrieving nearest neighbours in the codebook-quantized embedding space. Recall@$k$ ($k \in \{1, 5, 10\}$), which measures whether the correct match appears within the top-$k$ retrieved candidates, is reported.

\noindent\textbf{4. Cross-modal Video Segmentation (AVSBench-S4)~\cite{AVS}:}  
Using the AVT setup, we train a query-based video segmenter with one modality (e.g., text) and directly test segmentation performance in another modality (e.g., audio). We report mean Intersection-over-Union (mIoU) and F1 score.

Precision is used for single-label datasets (AVE), while segment-level F1 and accuracy are reported for multi-label or localization tasks (AVVP, AVSBench).

Additional cross-dataset transfer experiments are provided in the supplementary material.

%\noindent\textbf{Feature backbones:} %Following AVE~\cite{AVE},
%We use VGG19~\cite{VGG19} for visual feature extraction, a VGG-like CNN~\cite{VGGlike} as the audio feature backbone, and BERT~\cite{BERT} for text features.%, following Xia et al.~\cite{xia2023achieving}.

\subsection{Implementation Details}

We use the unimodal discretization framework from \cref{subsubsec:baseline} as our baseline and additionally compare CoDAAR with state-of-the-art multimodal discrete representation and domain generalization methods: CMCM~\cite{liu2021cross}, CODIS~\cite{CODIS}, TURN~\cite{TURN}, DCID~\cite{xia2023achieving}, and MICU~\cite{huang2025opensetcrossmodalgeneralization}. Since CMCM, CODIS, and TURN primarily align only \emph{two} modalities and cannot reliably generalize to unconstrained tri-modal setups, we compare AVT settings only with DCID and MICU, which support tri-modal alignment. All experiments use embedding dimension \(D{=}256\) and \(K{=}800\) codewords per modality. Temporal EMA updates use \(\lambda_{\text{self}}{=}0.6\), \(\lambda_{\text{cross}}{=}0.2\) for tri-modal (AVT) and \(\lambda_{\text{self}}{=}0.75\), \(\lambda_{\text{cross}}{=}0.25\) for bi-modal (AV). Complete implementation details can be found in the supplementary material.

\subsection{Comparison with SOTA}
\textbf{Cross-modal Event Classification and Localization:}  
Tables~\ref{tab:sota_ave_avvp} and~\ref{tab:sota_avt_ave_avvp} report results for the AV and AVT pretraining settings on AVE (global classification) and AVVP (segment-level localization).  
Without cross-modal alignment, the unimodal baseline transfers knowledge poorly between modalities. Among prior methods, DCID performs well for smaller datasets, while MICU yields stronger classification but weaker localization. CoDAAR achieves the most balanced and consistent results, particularly on AVVP. While DCID and MICU's unified codebooks produce strictly modal-agnostic representations, our discrete representations retain both modality-specific cues and modality-agnostic semantics, enabling fine-grained localized multi-event segment-level decisions. In the larger 90K-scale setting, CoDAAR maintains stable accuracy, whereas DCID and MICU degrade, suggesting that our temporal–hierarchical alignment helps mitigate overfitting from instances with noisy multimodal alignment in larger datasets.  
Incorporating text during pretraining (AVT) further improves cross-modal alignment. Text event descriptions act as holistic semantic anchors, stabilizing codebook index synchronization. While MICU and DCID achieve slightly higher AVE precision in small-scale setups, CoDAAR outperforms on the multi-event AVVP benchmark and remains robust across scales.

\noindent\textbf{Cross-Modal Zero-Shot Retrieval:}
Table~\ref{tab:sota_retrieval_main} reports zero-shot retrieval on MSCOCO (V$\leftrightarrow$T) and Clotho (A$\leftrightarrow$T). CoDAAR achieves the highest overall average across both scales, with strong gains on Clotho under both 40K and 90K settings, while remaining competitive on MSCOCO at lower thresholds and the 40K setting and leading at R@10 and under 90K. This confirms generalization to unseen datasets and modality pairs without fine-tuning.

\begin{figure}
\resizebox{0.85\columnwidth}{!}{%
    \centering
    \includegraphics[width=1\linewidth]{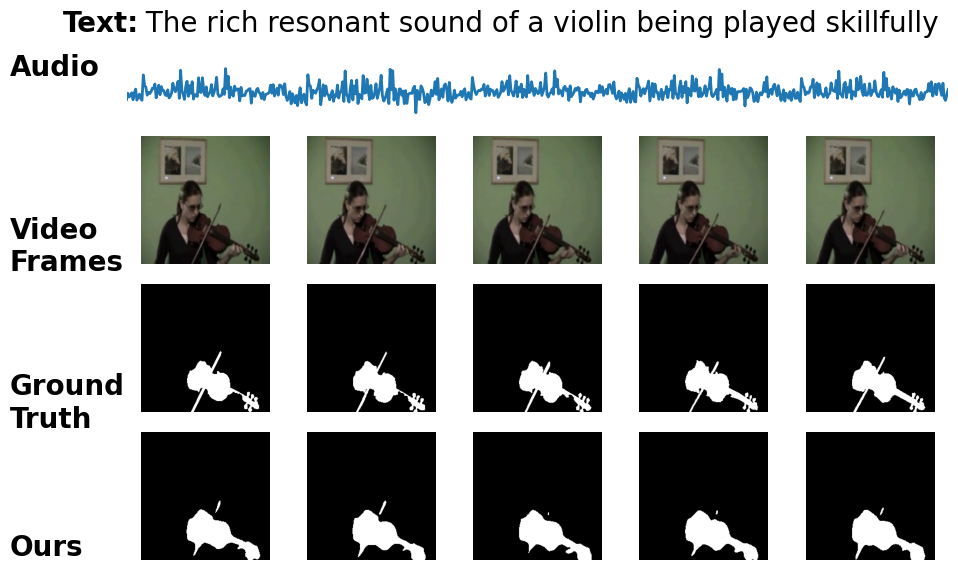}
    }
    \caption{Visualization of audio-to-text generalization on AVS-S4 video segmentation task}
    \label{fig:A2T}
    \vspace{-10pt}
\end{figure}

\begin{figure}
\resizebox{0.85\columnwidth}{!}{%
    \centering
    \includegraphics[width=1\linewidth]{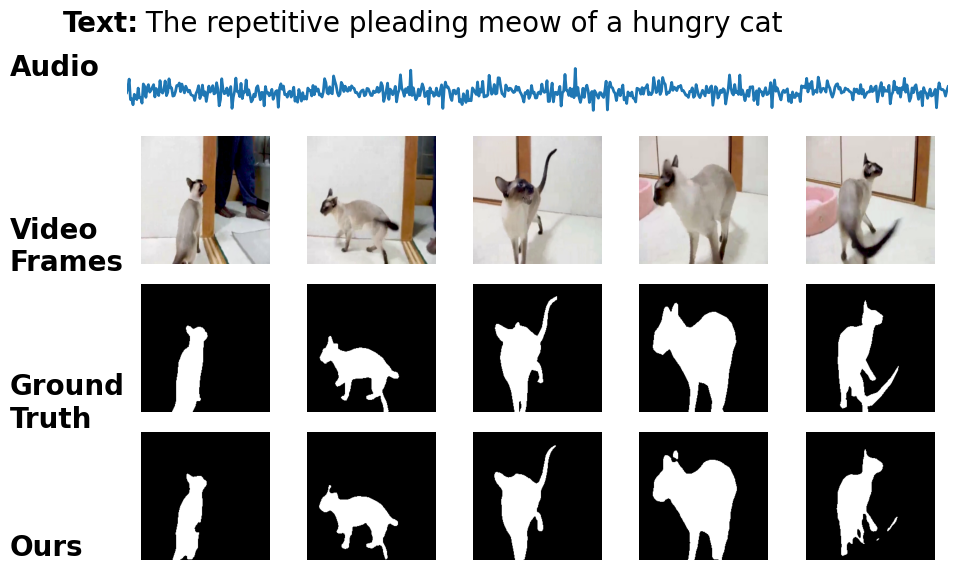}
    }
    \caption{Visualization of text-to-audio generalization on AVS-S4  video segmentation task.}
    \label{fig:T2A}
    \vspace{-10pt}
\end{figure}

\noindent\textbf{Cross-modal Video Segmentation:}  
Table~\ref{tab:avs_s4} presents results on the AVSBench-S4 benchmark. CoDAAR attains the best or comparable mIoU and F1 scores in both directions (audio $\!\leftrightarrow\!$ text).  
The performance advantage is attributed to the tri-modal centroids, which retain richer visual semantics, thereby improving segmentation even when queries are issued from a different modality. Qualitative maps (\cref{fig:A2T,fig:T2A}) further confirm precise visual localization, demonstrating CoDAAR’s ability to generalize to unseen modality combinations. Additional qualitative visualizations can be found in the supplementary material.

Overall, CoDAAR provides consistent improvements across tasks and datasets, resulting in stable cross-modal and cross-domain generalization.

\subsection{Ablation Studies}\label{subsec:ablation}
\begin{table}[t]
  \centering
  \setlength{\tabcolsep}{4pt}
  \renewcommand{\arraystretch}{1.08}
  \resizebox{0.6\columnwidth}{!}{%
    \begin{tabular}{cccc|cc|cc}
    \hline
    \multicolumn{4}{c|}{\textbf{Components}} & \multicolumn{2}{c|}{\textbf{AVE}} & \multicolumn{2}{c}{\textbf{AVVP}} \\
    \cline{1-8}
    CPC & DTA & CSA & DQA & V$\!\to\!$A & A$\!\to\!$V & V$\!\to\!$A & A$\!\to\!$V \\
    \hline
     --  & \checkmark & \checkmark & \checkmark & 38.6 & 40.2 & 52.5 & 53.1 \\ % w/o CPC
    \checkmark &    --      &  --     & \checkmark & 19.6 & 9.7 & 7.3 & 8.3 \\ % w/o DTA + CSA
    \checkmark & \checkmark   &  --     & \checkmark & 21.2 & 18.1 & 10.4 & 11.2 \\ % w/o CSA
    \checkmark &    --    &  \checkmark   & \checkmark & 45.2 & 47.3 & 60.5 & 69.0 \\ % w/o DTA
    \checkmark & \checkmark & \checkmark &   --       & 43.5 & 49.5 & 65.9 & 68.1 \\ % w/o DQA
    \hline
    \checkmark & \checkmark & \checkmark & \checkmark & \textbf{47.6} & \textbf{49.7} & \textbf{63.6} & \textbf{72.4} \\ % Full
    \hline
    \multicolumn{8}{c}{\emph{Full model}} \\
    \hline
  \end{tabular}}
   \caption{Component Ablation (AV settings): AVE (Precision), AVVP (Segment Accuracy)}
  \label{tab:ablate_av_40k_ticks}
\end{table}

\begin{table}[t]
  \centering
  \setlength{\tabcolsep}{4pt}
  \renewcommand{\arraystretch}{1.08}
  \resizebox{0.6\columnwidth}{!}{%
  \begin{tabular}{cccc|cc|cc}
    \hline
    \multicolumn{4}{c|}{\textbf{Components}} & \multicolumn{2}{c|}{\textbf{AVE}} & \multicolumn{2}{c}{\textbf{AVVP}} \\
    \cline{1-8}
    CPC & DTA & CSA & DQA & V$\!\to\!$A & A$\!\to\!$V & V$\!\to\!$A & A$\!\to\!$V \\
    \hline
     --  & \checkmark & \checkmark & \checkmark & 42.6 & 42.9 & 48.1 & 56.7 \\ % w/o CPC
    \checkmark &    --      &  --     & \checkmark & 3.4 & 6.9 & 4.8 & 10.8 \\ % w/o DTA + CSA
    \checkmark & \checkmark   &  --     & \checkmark & 11.1 & 25.5 & 10.7 & 7.5 \\ % w/o CSA
    \checkmark &    --    &  \checkmark   & \checkmark & 46.9 & 51.5 & 66.3 & 70.6 \\ % w/o DTA
    \checkmark & \checkmark & \checkmark &   --       & 51.5 & 54.8 & 69.5 & 71.0 \\ % w/o DQA
    \hline
    \checkmark & \checkmark & \checkmark & \checkmark & \textbf{52.3} & \textbf{55.5} & \textbf{70.8} & \textbf{72.5} \\ % Full
    \hline
    \multicolumn{8}{c}{\emph{Full model}} \\
    \hline
  \end{tabular}}
  \caption{Component Ablation (AVT settings): AVE (Precision), AVVP (Segment Accuracy)}
  \label{tab:ablate_avt_40k_ticks}
\end{table}

% ---------- Cascading Order Ablation (single-column, no averages) ----------
% \begin{table}[t]
%   \centering
%   \setlength{\tabcolsep}{5pt}
%   \renewcommand{\arraystretch}{1.12}
%   \resizebox{0.6\columnwidth}{!}{%
%   \begin{tabular}{l|l|cc|cc}
%     \hline
%     \multirow{2}{*}{Setting} & \multirow{2}{*}{Cascade}
%       & \multicolumn{2}{c|}{\textbf{AVE}}
%       & \multicolumn{2}{c}{\textbf{AVVP}} \\
%     \cline{3-6}
%       &  & V$\!\to$A & A$\!\to$V & V$\!\to$A & A$\!\to$V \\
%     \hline
%     AV   & V$\!\rightarrow$A & 46.5 & \textbf{50.0} & 62.6 & 70.3 \\
%          & A$\!\rightarrow$V & \textbf{47.6} & 49.7 & \textbf{63.6} & \textbf{72.4} \\
%     \hline
%     AVT  & T$\!\rightarrow$V$\!\rightarrow$A & 51.4 & 54.1 & 69.5 & 71.0 \\
%          & T$\!\rightarrow$A$\!\rightarrow$V & \textbf{52.3} & \textbf{55.5} & \textbf{70.8} & \textbf{72.5} \\
%     \hline
%   \end{tabular}
%   }
%   \caption{Cascading order ablation for AV and AVT on AVE (Precision) and AVVP (Segment-level Accuracy).}
%   \label{tab:cascade_ablation}
%   \vspace{-7pt}
% \end{table}

\begin{table}[t]
  \centering
  \setlength{\tabcolsep}{5pt}
  \renewcommand{\arraystretch}{1.12}
  \resizebox{0.6\columnwidth}{!}{%
  \begin{tabular}{l|l|cc|cc}
    \hline
    \multirow{2}{*}{Setting} & \multirow{2}{*}{Cascade}
      & \multicolumn{2}{c|}{\textbf{AVE}}
      & \multicolumn{2}{c}{\textbf{AVVP}} \\
    \cline{3-6}
      &  & V$\!\to$A & A$\!\to$V & V$\!\to$A & A$\!\to$V \\
    \hline
    AV   & V$\!\rightarrow$A & 46.5 & \textbf{50.0} & 62.6 & 70.3 \\
         & A$\!\rightarrow$V & \textbf{47.6} & 49.7 & \textbf{63.6} & \textbf{72.4} \\
    \hline
    AVT  & T$\!\rightarrow$A$\!\rightarrow$V & \textbf{52.3} & \textbf{55.5} & \textbf{70.8} & 72.5 \\
         & T$\!\rightarrow$V$\!\rightarrow$A & 51.4 & 54.1 & 69.5 & 71.0 \\
         & A$\!\rightarrow$T$\!\rightarrow$V & 51.6 & 53.6 & 68.5 & 71.9 \\
         & A$\!\rightarrow$V$\!\rightarrow$T & 51.2 & 52.3 & 70.5 & \textbf{73.0} \\
         & V$\!\rightarrow$T$\!\rightarrow$A & 51.4 & 53.8 & 70.0 & 71.4 \\
         & V$\!\rightarrow$A$\!\rightarrow$T & 50.5 & 52.2 & 70.1 & 72.3 \\
    \hline
  \end{tabular}
  }
  \caption{Cascading order ablation on AVE (Precision) and AVVP (Segment-level Accuracy). AV uses VGGSound-AVEL 40K; AVT exhaustively ablates all six permutations.}
  \label{tab:cascade_ablation}
  % \vspace{-7pt}
\end{table}

All ablation experiments are conducted on the VGGSound-AVEL 40K split to ensure consistency across analyses.  
\noindent\textbf{Component Analysis:}  
Tables~\ref{tab:ablate_av_40k_ticks} and~\ref{tab:ablate_avt_40k_ticks} report the impact of removing each module under AV and AVT settings.  
Removing \emph{CSA} leads to the most severe performance degradation (e.g., AVE $21.2/18.1$, AVVP $10.4/11.2$ in AV), confirming that cross-modal semantic alignment of modality-specific codebooks is essential for index consensus.  
Eliminating \emph{CPC} also causes a major drop (AVVP $63.6/72.4 \rightarrow 52.5/53.1$ in AV; $70.8/72.5 \rightarrow 48.1/56.7$ in AVT), as this module injects cross-modal cues into unimodal encoders.  The absence of \emph{DTA} or \emph{DQA} individually weakens performance—particularly classification precision when DQA is removed ($52.3/55.5 \rightarrow 51.5/54.8$ in AVT)—indicating that temporal and quantization alignment are complementary.  Overall, the full model achieves the best performance in all transfer directions, demonstrating the importance of their joint interaction.

\noindent\textbf{Cascade Order:}  
Table~\ref{tab:cascade_ablation} ablates the hierarchical consensus order. In AV, A$\!\rightarrow$V outperforms V$\!\rightarrow$A, suggesting that mapping semantics closer to video codewords benefits from video's richer spatial detail. For AVT, we exhaustively evaluate all six permutations: T$\!\rightarrow$A$\!\rightarrow$V is optimal, as text first provides holistic semantic anchoring while video last refines alignment with spatial cues. Orderings placing text last yield lower AVE scores, confirming its role as an early semantic anchor.

Further ablations on codebook size and embedding dimension are included in the supplementary material.

% (AV: A$\!\rightarrow$V $>$ V$\!\rightarrow$A; AVT: T$\!\rightarrow$A$\!\rightarrow$V $>$ T$\!\rightarrow$V$\!\rightarrow$A).  

\section{Conclusion}
In this paper, we presented \textbf{CoDAAR}, a cross-modal discrete alignment module that introduces two novel mechanisms—DTA and CSA—to synchronize modality-specific codebooks at the index level semantically. These modules jointly preserve modality-specific and modality-agnostic information within a unified discrete space. Future work includes expanding CoDAAR to new modalities (e.g., sensor streams, point clouds) and applications such as sentiment analysis and cross-modal retrieval. 

\section{Acknowledgements}
This work was supported by the Federal Ministry of Research, Technology and Space of Germany [project name: EMuLE – Enhancing Data and Model Efficiency in Multimodal Learning; grant number 16IS24059]. The last author was partially funded by the Lower Saxony Ministry of Science and Culture (MWK) with funds from the Volkswagen Foundation’s Zukunft Niedersachsen program [project name: CAIMed - Lower Saxony Center for Artificial Intelligence and Causal Methods in Medicine; grant number: ZN4257]. The authors acknowledge the Hannover Medical School for providing MHH-HPC resources that have contributed to the results reported in this paper.
{
    \small
    \bibliographystyle{ieeenat_fullname}
    \bibliography{main}
}
\clearpage
\setcounter{page}{1}
\maketitlesupplementary

\section{Notation Library}\label{app:notation}

\noindent\textbf{Conventions:}
$[\cdot;\cdot]$: channel-wise concatenation;
$\|\cdot\|_2$: $\ell_2$ norm;
$\mathrm{sg}[\cdot]$: stop–gradient operator;
$\mathbb{1}[\cdot]$: indicator function;

\paragraph{Modalities, indices, sizes:}
$m\in\{a,v,t\}$: modality (audio, video, text);
$m_1\neq m_2$: CMG train/test modalities;
$i$: sample index;
$t\in\{1{:}T\}$: time index;
$N$: number of samples;
$T$: timesteps per sample;
$D$: embedding dimensionality;
$K$: number of codewords per codebook;
$B$: batch size;
$H$: CPC prediction horizon;

\paragraph{Data and encoders:}
$\mathbf{X}=\{(x_i^a,x_i^v,x_i^t)\}_{i=1}^{N}$: paired multimodal dataset;
$\mathbf{x}_i^{m}$: input of modality $m$ for sample $i$;
$\mathbf{y}_i^{m}$: label of modality $m$ for sample $i$;
$\Psi^m(\cdot)$: modality-specific encoder;
$z_i^{m}=\Psi^m(x_i^{m})\in\mathbb{R}^{T\times D}$: encoder embedding sequence;
$z_{i,t}^{m}\in\mathbb{R}^{D}$: encoder embedding frame at timestep $t$;

\paragraph{Quantization and codebooks:}
$\mathrm{VQ}(\cdot)$: vector quantizer;
$\mathbf{E}_m=\{\mathbf{e}_m(k)\}_{k=1}^{K}\in\mathbb{R}^{K\times D}$: codebook of modality $m$;
$\mathbf{e}_m(k)\in\mathbb{R}^{D}$: codeword $k$ of modality $m$;
$k_{i,t}=\arg\min_{j}\|z_{i,t}^{m}-\mathbf{e}_m(j)\|_2^2$: nearest-neighbour codeword index;
$\hat{z}_{i,t}^{m}=\mathrm{VQ}(z_{i,t}^{m})=\mathbf{e}_m(k_{i,t})$: quantized latent at timestep $t$;
$\mathbf{E}=[\mathbf{E}_a;\mathbf{E}_v;\mathbf{E}_t]$: concatenated unified vocabulary;

\paragraph{Decoders, projections, heads:}
$P_m(\cdot)$: modality-specific projection;
$\mathcal{D}_m(\cdot)$: reconstruction decoder for modality $m$;
$\mathcal{G}(\cdot)$: modality-invariant downstream task head;
$\mathcal{E_L}(\cdot)$: downstream evaluation loss;

\paragraph{CPC (cross-modal information maximization):}
$(a,b)$: ordered modality pair (predict $b$ from $a$);
$c_t^{a}$: context from uniLSTM over modality $a$ at time $t$;
$W_h^{a}$: step-specific linear map at horizon step $h$ for modality $a$;
$z^{b}_{t+h}$: future latent of modality $b$ at offset $h$;
$\{z_j^{b}\}$: negatives drawn from modality $b$;
$H$: number of future steps predicted;

\paragraph{DQA (Discrete Quantization Alignment):}
$p_m^{\,i}\in\mathbb{R}^{K}$: sequence-level code-usage distribution for sample embedding $z_i^m$ and modality $m$;
$\mathcal{L}_{\mathrm{CMCM}}$ \cite{liu2021cross}: cross-modal code-matching loss aligning $\{p_m^{\,i}\}$ across modalities;

\begin{equation} \mathcal{L}_{\text{CMCM}} = -\frac{1}{N}\sum_{i}\! \log\! \frac{\exp\!\big(\langle p_a^{\,i}, \log p_b^{\,i}\rangle + \langle p_b^{\,i}, \log p_a^{\,i}\rangle\big)} {\sum_{j}\exp\!\big(\langle p_a^{\,i}, \log p_b^{\,j}\rangle + \langle p_b^{\,j}, \log p_a^{\,i}\rangle\big)}, \label{eq:cmcm} \end{equation} 
where \(p_m^{\,i}\!\in\!\Delta^{K-1}\) is the time-averaged soft assignment over indices \(k=1..K\) for modality \(m\) and embedding sample \(i\).

\paragraph{DTA (Discrete Temporal Alignment with EMA):}
$\pi_m(i,t)$: hard nearest-neighbour codeword index for $(i,t)$ in modality $m$;
$q_{m,i,t}(k)\in\{0,1\}$: one-hot responsibility that codeword $k$ was selected at $(i,t)$;
$\{n,p\}=\{a,v,t\}\setminus\{m\}$: the two modalities other than $m$;
$\mathbf{U}_m^{\mathrm{self}}(k)$: accumulator of modality-$m$ features assigned to $k$;
$\mathbf{U}_m^{\mathrm{cross}}(k)$: time-aligned accumulator of $(n,p)$ features assigned to $k$;
$\lambda_{\mathrm{self}}=0.6$: self-term mixing weight;
$\lambda_{\mathrm{cross}}=0.2$: cross-term mixing weight (per other modality);
$\rho\in(0,1)$: EMA decay factor;
$\varepsilon>0$: numerical stabilizer;
$\mathbf{H}_m(k)$: mixed accumulator for $k$ (self + cross);
$\mathbf{S}_m(k)$: EMA-weighted sum of assigned features for $k$;
$C_m(k)$: EMA-weighted assignment count for $k$;
$\mathbf{S}_m^{\mathrm{new}}(k)$, $C_m^{\mathrm{new}}(k)$: updated EMA statistics for $k$;
$\mathbf{e}_m^{\mathrm{new}}(k)$: updated centroid/codeword at index $k$;

\paragraph{CSA (Cascading Semantic Alignment):}
$\mathbf{e}_v^0(k)$, $\mathbf{e}_a^0(k)$, $\mathbf{e}_t^0(k)$: post-DTA (pre-cascade) centroids for video, audio, text at index $k$;
$\mathbf{c}^0(k)=\tfrac{1}{3}(\mathbf{e}_v^0(k)+\mathbf{e}_a^0(k)+\mathbf{e}_t^0(k))$: cross-modal semantic anchor at index $k$;
$\mathbf{e}_t^1(k)$, $\mathbf{e}_a^1(k)$, $\mathbf{e}_v^1(k)$: centroids after cascade (applied in order $t\!\rightarrow\!a\!\rightarrow\!v$);
$^0$, $^1$: superscripts denoting pre-/post-cascade values within the same iteration;
$k$: codeword index shared across modalities;

\paragraph{Reconstruction:}
$\hat{z}_i^{m}\in\mathbb{R}^{T\times D}$: quantized sequence for modality $m$;
$\hat{z}_i^{\mathrm{tri}}=[\hat{z}_i^{a};\hat{z}_i^{v};\hat{z}_i^{t}]\in\mathbb{R}^{T\times 3D}$: tri-modal quantized code;
$\tilde{\mathbf{x}}^{\,m}_i=\mathcal{D}_m([\hat{z}_i^{\mathrm{tri}};P_m(\mathbf{x}_i^{m})])$: reconstruction of modality $m$ for sample $i$;

\paragraph{Other:}
$N_{\mathrm{re}}$: reset threshold for dead codes (consecutive batches unselected);

\section{Implementation Details}\label{app:impl}

\subsection{Pretraining Setup}\label{app:impl:pretrain}
\paragraph{\textit{Backbone features:}}
Following \cite{AVE}, for every 1\,s video segment, we sample 16 RGB frames and extract pool5 activations from a VGG-19 model \cite{VGG19}. The 16 frame-wise tensors are averaged using global average pooling to yield a $7\times 7\times 512-D$, $512= D_v$, visual descriptor per second. Audio is encoded at 1\,s granularity with a VGG-style network pretrained on AudioSet, producing $128-D = D_a$ features \cite{VGGlike} for every second. For text, we use a short descriptive sentence for event labels using handwritten templates from \cite{xia2023achieving} and encode tokens with BERT to obtain $768-D$ word features \cite{BERT}. Then we use a bidirectional LSTM to temporally contextualize the tokens, yielding $256-D = D_t$ text context vectors.

\subsubsection*{\textit{\textbf{Modal-Specific Encoders:}}}

\paragraph{Video Encoder $\Psi^v(\cdot)$:} 
Following the architecture from \cite{Xu2020CMRAN}, we employ a two-stage video encoding process. First, a channel-spatial attention mechanism processes the input video features $\mathbf{x}^v \in \mathbb{R}^{B \times T \times H \times W \times D_v}$ to produce spatially-aggregated features $\mathbf{g}^v_{att} \in \mathbb{R}^{B \times T \times D_v}$ while simultaneously preserving spatial information through convolutional transformations, yielding $\mathbf{g}^v_{spatial} \in \mathbb{R}^{B \times T \times H' \times W' \times D'_v}$. The aggregated features $\mathbf{g}^v_{att}$ are first projected to the model dimension $D=256$ through linear transformations, then processed through an Internal Temporal Relation Module consisting of a 2-layer transformer encoder with 4-head self-attention to capture temporal dependencies, producing the final video representation $\mathbf{z}^v \in \mathbb{R}^{B \times T \times D}$.

\paragraph{Audio Encoder $\Psi^a(\cdot)$ and Text Encoder $\Psi^t(\cdot)$:}
Both audio and text modalities employ the Internal Temporal Relation Module directly. The audio features $\mathbf{x}^a \in \mathbb{R}^{B \times T \times D_a}$ and text features $\mathbf{x}^t \in \mathbb{R}^{B \times T \times D_t}$ are first projected to the model dimension $D=256$ through linear transformations, then processed through the same 2-layer transformer architecture seperately to capture temporal relationships, yielding $\mathbf{z}^a \in \mathbb{R}^{B \times T \times D}$ and $\mathbf{z}^t \in \mathbb{R}^{B \times T \times D}$ respectively.

\subsubsection*{\textit{\textbf{Modality-Specific Projections and Decoders:}}}

\paragraph{Projection Layers:}
Due to the inherent spatial complexity of visual data, we employ distinct projection strategies for each modality input features $x^m,m\in\{a,v,t\} $. For video, we leverage the spatially preserved features $\mathbf{g}^v_{spatial}$ from the encoder's intermediate stage, which serve as $P_v(\cdot)$ and maintain the spatial structure necessary for accurate reconstruction. For audio and text modalities, we apply simple linear projections $P_a(\cdot): \mathbb{R}^{D_a} \rightarrow \mathbb{R}^{D}$ and $P_t(\cdot): \mathbb{R}^{D_t} \rightarrow \mathbb{R}^{D}$ respectively, as these modalities lack the spatial dimensionality that requires preservation.

\vspace{-3pt}

\paragraph{Video Decoder $\mathcal{D}^v(\cdot)$:}
The video decoder addresses the spatial reconstruction challenge through a convolutional architecture. The trimodal codes for every batch, $\hat{\mathbf{z}}^{\mathrm{tri}}_i = [\hat{\mathbf{z}}^{a}_i; \hat{\mathbf{z}}^{v}_i; \hat{\mathbf{z}}^{t}_i] \in \mathbb{R}^{B\times T\times 3D}$, are first projected and spatially broadcasted to match the spatial dimensions of $\mathbf{h}^v_{spatial}$. After concatenation along the channel dimension, a series of transposed convolutions with residual connections progressively upsample the features: $\hat{\mathbf{x}}^v = \text{ConvTranspose}([\text{Broadcast}(\hat{\mathbf{z}}^{\mathrm{tri}}); \mathbf{h}^v_{spatial}])$, where the decoder reconstructs the original spatial resolution $\mathbb{R}^{B \times T \times H \times W \times D_v}$.

\paragraph{Audio Decoder $\mathcal{D}^a(\cdot)$ and Text Decoder $\mathcal{D}^t(\cdot)$:}
Both audio and text decoders employ an identical architecture to reconstruct their respective modality features from the batch-wise trimodal representations $\hat{\mathbf{z}}^{\mathrm{tri}}_i = [\hat{\mathbf{z}}^{a}_i; \hat{\mathbf{z}}^{v}_i; \hat{\mathbf{z}}^{t}_i] \in \mathbb{R}^{B\times T\times 3D}$. For audio reconstruction, the trimodal codes are first projected through two successive linear transformations: $\hat{\mathbf{z}}^a_{proj} = \text{Linear}(\text{Linear}(\hat{\mathbf{z}}^{\mathrm{tri}}))$, then concatenated with the audio projection features $P_a(\mathbf{x}^a)$ and passed through a final linear layer to reconstruct the original audio: $\hat{\mathbf{x}}^a = \text{Linear}([\hat{\mathbf{z}}^a_{proj}; P_a(\mathbf{x}^a)]) \in \mathbb{R}^{B \times T \times D_a}$. Similarly, the text decoder processes the trimodal codes through $\hat{\mathbf{z}}^t_{proj} = \text{Linear}(\text{Linear}(\hat{\mathbf{z}}^{\mathrm{tri}}))$ and combines them with the text projection features to reconstruct: $\hat{\mathbf{x}}^t = \text{Linear}([\hat{\mathbf{z}}^t_{proj}; P_t(\mathbf{x}^t)]) \in \mathbb{R}^{B \times T \times D_t}$, where $P_t(\cdot)$ and $P_a(\cdot)$ represent the text-specific and audio-specific projection.

\paragraph{\textit{Cross-modal prediction CPC:}}
We employ Cross-CPC between modality pairs to inject fine-grained temporal cross-modal signals \cite{DBLP:journals/corr/abs-1807-03748}. The prediction horizon is set to $H{=}2$ for both the audio–visual stage and the audio–visual–text (trimodal) stage, and we use a single-layer unidirectional LSTM as our autoregressive model to contextualize past embedding frames of each modality.

\paragraph{\textit{Optimization:}}
We pre-train for $6$ epochs with Adam at a learning rate of $4{\times}10^{-4}$, batch size 80, and EMA decay $\rho{=}0.99$ for codebook updates. Inactive entries are reset when they have not been selected for $N_{\mathrm{re}}{=}300$ consecutive mini-batches. All pretraining runs are performed on a single NVIDIA A100 GPU.

\subsection{Downstream Tasks}\label{app:impl:downstream}
Across all tasks, encoders and codebooks are \textit{frozen}. Only lightweight task heads are trained on the trimodal codebook. All downstream experiments run on a single NVIDIA A100 GPU. 

\paragraph{\textit{Cross-modal event classification (AVE):}}
AVE contains 28 classes with 10\,s audio–video clips \cite{AVE}. Using the pretrained encoder, a 10\,s video produces a sequence of 10 trimodal discrete vectors. A two-layer MLP maps these to a 28-D class space with softmax and cross-entropy. We use a learning rate of $2.5{\times}10^{-4}$ and a batch size of 256. After training on one source modality (e.g., V$\rightarrow$A), we swap the input modality (A) at evaluation time and reuse the same head to evaluate zero-shot transfer; A$\rightarrow$V is symmetric.

\begin{figure*}[t]
  \centering
  \includegraphics[height=0.33\textheight, keepaspectratio]{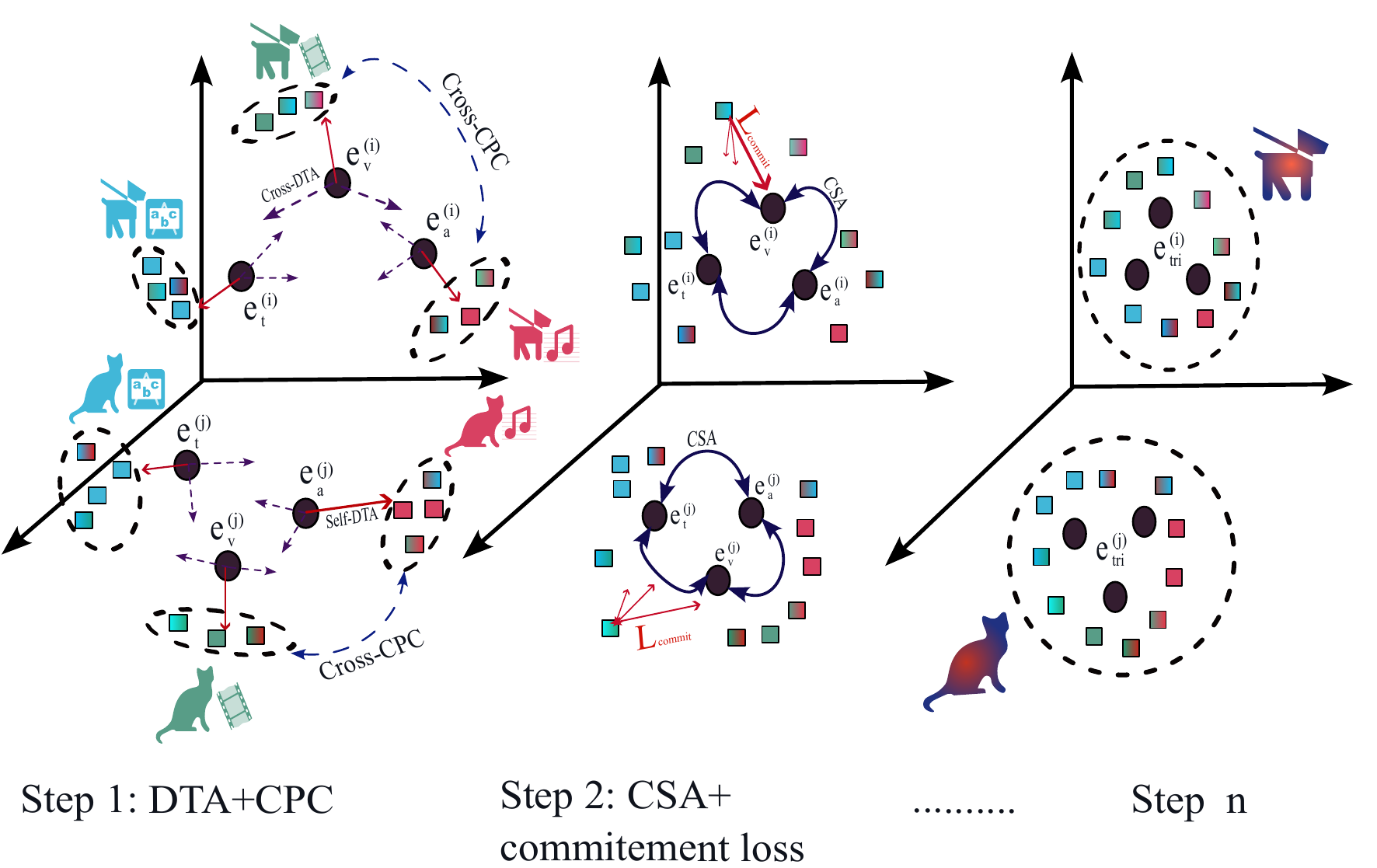}
  \caption{Cross-modal Discrete Alignment Framework}
  \label{fig:sub:seq_ema_1}
\end{figure*}

\paragraph{\textit{Cross-modal event localization (AVVP):}}
AVVP comprises 25 event types and 10\,s clips, but unlike AVE where each clip contains a single event type, AVVP segments can be multi-label \cite{AVVP}. Importantly, the audio and video modalities within the same clip may contain different event labels. For instance, in a single video segment, the audio track might contain events A and B, while the visual information contains events A, C, and D. We reuse the encoder and the trimodal codes and attach a three-layer MLP with sigmoid activations to produce 25 per-segment scores for multi-label classification; a standard multi-label cross-entropy loss is used to handle the presence of multiple simultaneous events. Hyperparameters match AVE.

\paragraph{\textit{Cross-modal video segmentation (AVS-S4):}}
AVS-S4 has 4{,}932 five-second videos over 23 categories \cite{AVS}. For A$\rightarrow$T, we encode audio into the unified discrete space using the  AVT settings pretrained encoder. The downstream segmentation task decoder follows the AVS architecture \cite{AVS} (visual backbone, audio–video interaction, and decoder). We use Adam with a learning rate of $1{\times}10^{-4}$ and batch size 4. After training, we replace the audio input with text to evaluate T$\rightarrow$A generalization (and vice versa for A$\rightarrow$T).

\paragraph{\textit{Cross-modal zero-shot retrieval (MSCOCO~\cite{lin2014microsoft}, Clotho~\cite{drossos2020clotho}):}}
MSCOCO contains 5{,}000 validation images with 5 captions each; Clotho provides 1{,}045 evaluation audio clips with 5 captions each. Visual features are extracted from VGG-19 pool5 activations and audio features from a VGGish model, using the same pipelines as pretraining. Captions are encoded with BERT and contextualized via a bidirectional LSTM. Each modality input is quantized through the frozen pretrained encoder and codebook, and the resulting discrete representations are temporally averaged to obtain a single embedding per sample. Retrieval is performed via nearest-neighbor lookup in the quantized embedding space without any task-specific fine-tuning.

\paragraph{\textit{Cross-modal \& cross-dataset localization (AVE $\!\rightarrow\!$ AVVP):}}
To measure generalization across both modality and dataset, we exploit the 12 shared classes between AVE and AVVP (e.g., \emph{dog}, \emph{car}, \emph{helicopter}, \emph{violin}, \emph{frying}, \emph{motorcycle}, \emph{acoustic guitar}, \emph{banjo}, \emph{baby cry}, \emph{chainsaw}, \emph{cat}, \emph{accordion}) \cite{AVE, AVVP}. We train a weakly supervised localization head, a two-layer MLP, on AVE (global events) using either video or audio and evaluate localization on the opposite modality in AVVP (local events) using the F1 score. Other settings mirror the AVVP localization protocol above. 

% \vspace{-10pt}

\paragraph{\textit{Cross-modal cross-dataset zero-shot transfer (UCF $\!\leftrightarrow\!$ VGGSound):}}
To evaluate cross-modal-dataset generalization, we conduct bidirectional zero-shot transfer between UCF-101 \cite{UCF} (video) and VGGSound \cite{CPSP} (audio) datasets, which share 16 common categories, including musical instruments (\emph{acoustic guitar, cello, flute, piano, sitar, tabla, violin}), sports activities (\emph{skiing, table tennis, frisbee, basketball, bowling, volleyball, skateboarding}), and other actions (\emph{typing on typewriter, rope skipping}). We train the model on one modality and evaluate on the other: UCF(v) $\!\rightarrow\!$ VGGSound(a) trains on video features and tests on audio, while VGGSound(a) $\!\rightarrow\!$ UCF(v) performs the reverse. We freeze the pretrained encoders and codebooks and attach a linear 2-layer MLP classifier to the quantized representations for 16-way classification. 

\section{Experiments Continued}
% ---------- AVT: AVE→AVVP & UCF(v)↔VGG(a) ----------
\begin{table*}[t]
  \centering
  \setlength{\tabcolsep}{6pt}
  \renewcommand{\arraystretch}{1.1}
  \resizebox{0.6\textwidth}{!}{%
  \begin{tabular}{l|cc|cc|cc|cc|c}
    \hline
    \multirow{2}{*}{Method}
      & \multicolumn{4}{c|}{\textbf{VGGSound-AVEL 40K}}
      & \multicolumn{4}{c|}{\textbf{VGGSound-AVEL 90K}}
      & \multicolumn{1}{c}{Average} \\
    \cline{2-9}
      & \multicolumn{2}{c|}{AVE $\!\to\!$ AVVP}
      & \multicolumn{2}{c|}{UCF(v) $\!\leftrightarrow\!$ VGG(a)}
      & \multicolumn{2}{c|}{AVE $\!\to\!$ AVVP}
      & \multicolumn{2}{c}{UCF(v) $\!\leftrightarrow\!$ VGG(a)}
      & \\
    \cline{2-9}
       & V$\rightarrow$A & A$\rightarrow$V & V$\rightarrow$A & A$\rightarrow$V
      & V$\rightarrow$A & A$\rightarrow$V & V$\rightarrow$A & A$\rightarrow$V & \\
    \hline
    Baseline & 1.5 & 4.3 & 17.1 & 12.3 & 2.5 & 3.3 & 19.2 & 14.4 & 9.3 \\
    DCID~\cite{xia2023achieving} & 53.0 & 52.4 & 67.1 & 60.6 & \textbf{57.8} & 57.5 & 69.9 & 66.5 & 60.6 \\
    MICU~\cite{huang2025opensetcrossmodalgeneralization} & \textbf{56.3} & 54.9 & \textbf{75.3} & 64.5 & 54.1 & 51.6 & \textbf{73.7} & \textbf{67.0} & 62.2 \\
    \textbf{CoDAAR (ours)} & 55.8 & \textbf{56.5} & 73.5 & \textbf{66.5} & 52.3 & \textbf{60.2} & 71.2 & 65.0 & \textbf{62.6} \\
    \hline
  \end{tabular}
  }
   \caption{AVT setting: comparison with state-of-the-art methods on (i) cross-dataset domain transfer AVE $\!\to\!$ AVVP (segment-level F1 score) and cross-dataset cross-modal classification UCF(v)$\leftrightarrow$VGG(a) (Precision).}
  \label{tab:sota_avt_transfer}
\end{table*}
\subsection{Cross-dataset domain transfer evaluation setup:}
To assess generalization across datasets and modalities, we further conduct two additional evaluations (Table~\ref{tab:sota_avt_transfer}): (i) train on AVE (global classification) on one modality and test zero-shot on AVVP (fine-grained localization) on the other modality, reporting segment-level F1; (ii) train a visual classifier on the 16-class subset of UCF101~\cite{UCF} and evaluate zero-shot audio classification on the corresponding 16-class subset of VGGSound-AVEL~\cite{CPSP}, and vice versa, reporting precision.

\subsection{Cross-dataset Domain Transfer Comparison with SOTA:}
\noindent\textbf{Cross-dataset Domain Transfer:}  
Table~\ref{tab:sota_avt_transfer} evaluates zero-shot transfer across datasets and modalities.  
On both AVE$\!\to\!$AVVP and UCF(v)$\!\leftrightarrow\!$VGG(a), CoDAAR achieves competitive or higher segment-level F1 and precision than DCID and MICU, reflecting strong cross-modal index reuse and improved representation sharing between visual and auditory domains.

\section{Ablations Continued}
\begin{figure}[t]
\resizebox{0.9\columnwidth}{!}{%
  \centering
  \includegraphics[width=\linewidth]{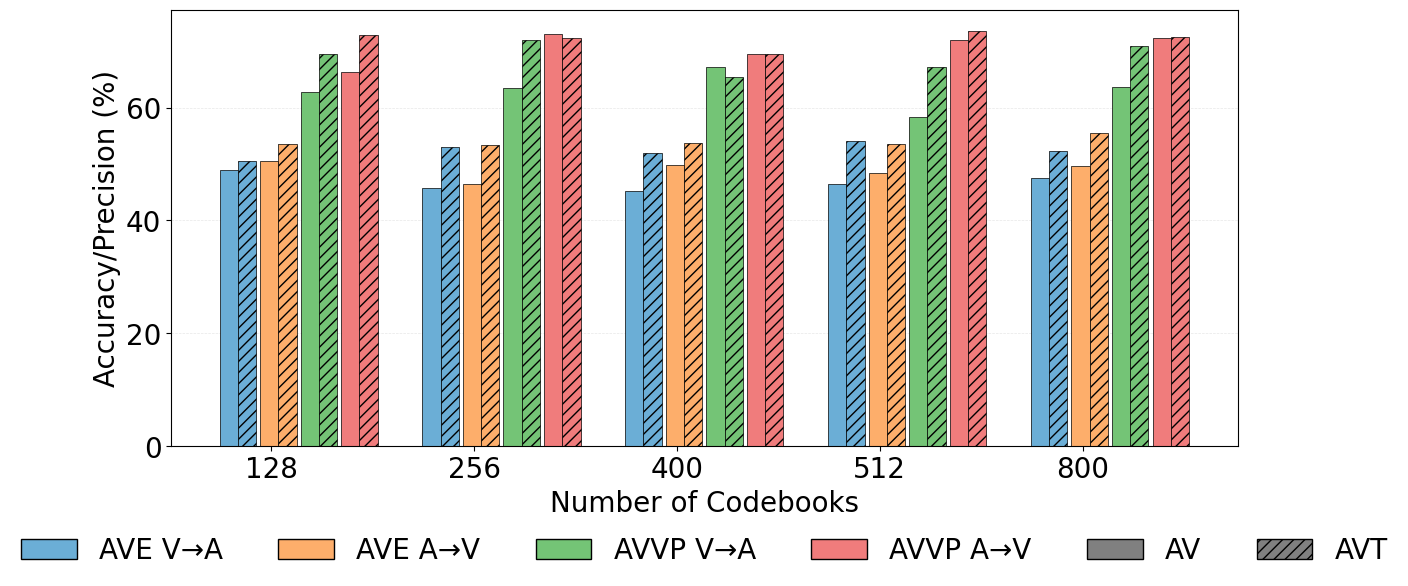}
  }
  \caption{Ablation on different codebook sizes in AV and AVT settings on two downstream tasks}
  \label{fig:codebook_ablation}
  % \vspace{-15pt}
\end{figure}
\begin{figure}[t]
\resizebox{0.9\columnwidth}{!}{%
  \centering
  \includegraphics[width=\linewidth]{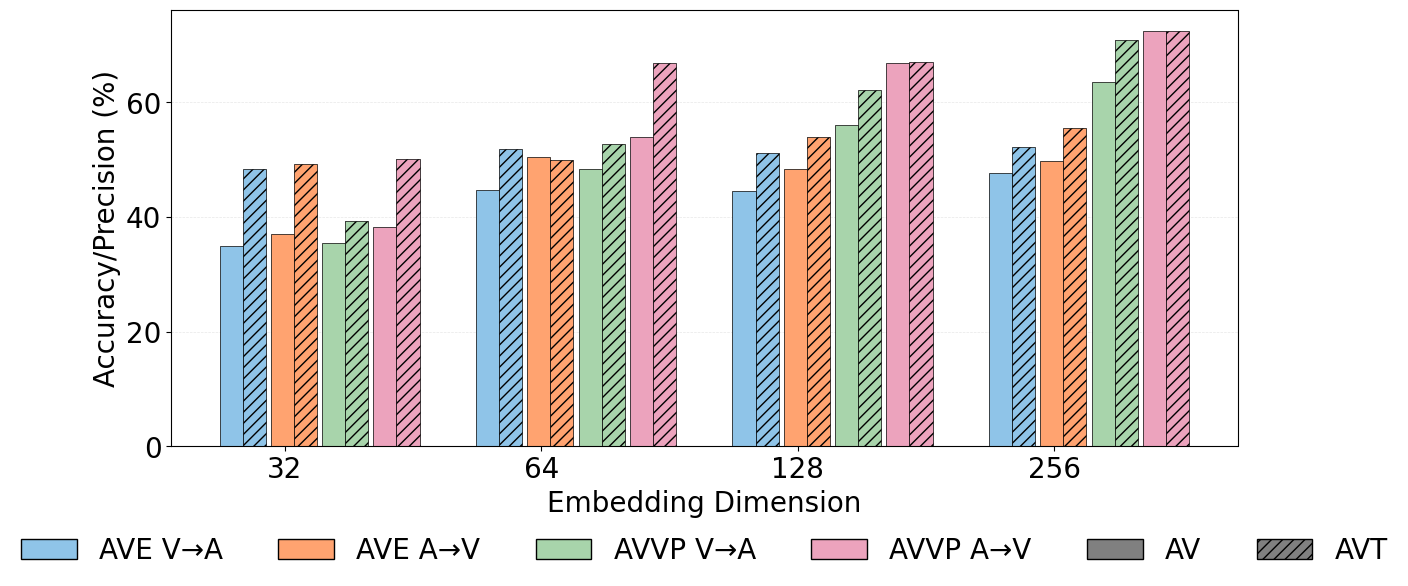}
  }
  \caption{Ablation on different dimension sizes in AV and AVT settings on two downstream tasks}
  \label{fig:dim_ablation}
\end{figure}
\subsection{Codebook Size Ablation}
Figure~\ref{fig:codebook_ablation} shows the effect of varying the number of codewords per modality.  Performance improves steadily from 128 to 800 entries in both settings, with the largest gains observed for AVVP. Larger codebooks allow finer multimodal quantization and better cross-modal index matching. We therefore adopt 800 entries per modality as a balanced trade-off between accuracy and efficiency.
\subsection{Embedding dimension Ablation}
Figure~\ref{fig:dim_ablation} shows a consistent trend: bigger dimensions enhance AVE and AVVP in all directions, with significant gains up to \textbf{256}. With more embedding dimensions, fine-grained details are captured by each encoder $\Psi^m$. Hence, we set the embedding dimension to 256 in all experiments.

\section{Computational Efficiency Comparison}\label{sec:efficiency}
Table~\ref{tab:efficiency_singlecol} compares computational efficiency and performance accuracy under the same audio-video-text pretraining setup on a single NVIDIA A100~(40\,GB) GPU. DCID~\cite{xia2023achieving} combines an additional CLUB-based information minimization objective with \emph{cross-attention-guided} EMA updates of a unified codebook, while MICU~\cite{huang2025opensetcrossmodalgeneralization} likewise uses cross-attention-guided EMA and adds further auxiliary constraints (e.g., a jigsaw loss). Consequently, both architectures increase per-epoch training time. In contrast, \textbf{CoDAAR} avoids heavy cross-attentional updates and extra objectives, using modality-specific codebooks with a mathematical alignment of DTA+CSA. Our modality-specific codebook design raises peak memory (three codebooks vs.\ one) but improves throughput (fewer minutes/epoch). Despite the higher memory footprint, CoDAAR attains the best downstream average on AVE classification and AVVP localization (61.7 vs.\ 58.2 for DCID and 51.6 for MICU), as shown in Table~\ref{tab:efficiency_singlecol} for the AVT setting.

\begin{table}[t]
  \centering
  \setlength{\tabcolsep}{1pt}
  \renewcommand{\arraystretch}{1.1}
  \resizebox{\columnwidth}{!}{%
  \begin{tabular}{l|cc|cc|c|c}
    \hline
    \multirow{2}{*}{Method} 
      & \multicolumn{2}{c|}{\textbf{VGGSound 40K}} 
      & \multicolumn{2}{c|}{\textbf{VGGSound 90K}} 
      & \multirow{2}{*}{Total Ep.} 
      & \multirow{2}{*}{Avg.} \\
    \cline{2-5}
      & GPU mem & Time/ep & GPU mem & Time/ep & & \\
    \hline
    DCID~\cite{xia2023achieving}  & 7.2  & 22.3 & 7.4  & 41.8 & 5 & 58.2 \\
    MICU~\cite{huang2025opensetcrossmodalgeneralization} & 8.4  & 25.6 & 8.5  & 54.5 & 5 & 51.6 \\
    \textbf{CoDAAR (ours)}         & 11.6 & 14.4 & 13.2 & 23.8 & 6 & \textbf{61.7} \\
    \hline
  \end{tabular}}
  \caption{Compute profile and downstream average (\textit{AVE+AVVP}) for each pretraining method in the AVT setting. 
  GPU mem is peak memory (GB); Time/ep is minutes per epoch. Total Ep. is the total number of pretraining epochs. 
  Measured on an NVIDIA A100~(40\,GB) GPU.}
  \label{tab:efficiency_singlecol}
\end{table}

\section{More AVS Generalization Visualization:}
\begin{figure}
\resizebox{0.9\columnwidth}{!}{%
    \centering
    \includegraphics[width=1\linewidth]{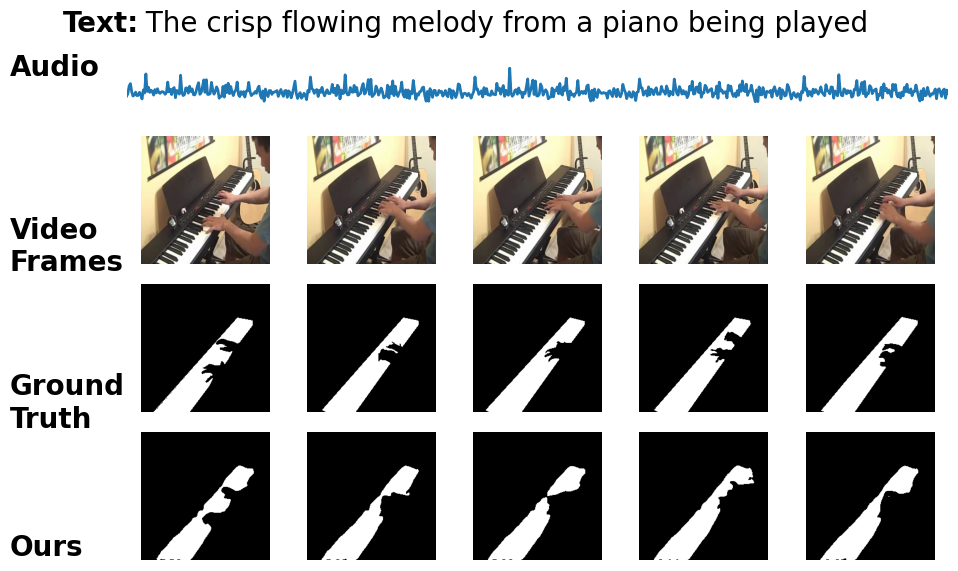}
    }
    \caption{Visualization of text-to-audio generalization on AVS-S4  video segmentation task - Piano playing}
    \label{fig:T2A_piano}
\end{figure}

\begin{figure}
\resizebox{0.9\columnwidth}{!}{%
    \centering
    \includegraphics[width=1\linewidth]{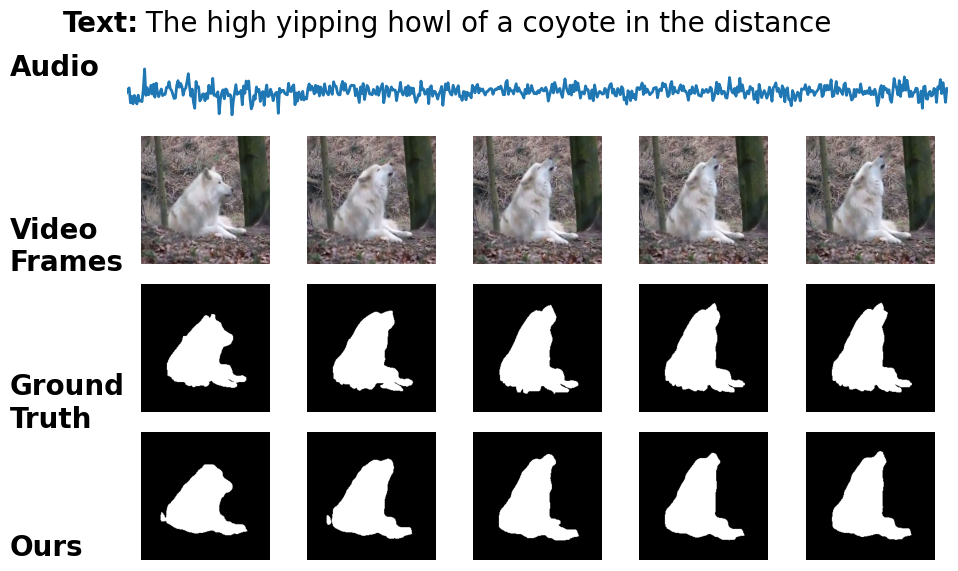}
    }
    \caption{Visualization of text-to-audio generalization on AVS-S4  video segmentation task – Coyote Howling}
    \label{fig:T2A_coyote}
\end{figure}

\begin{figure}
\resizebox{0.9\columnwidth}{!}{%
    \centering
    \includegraphics[width=1\linewidth]{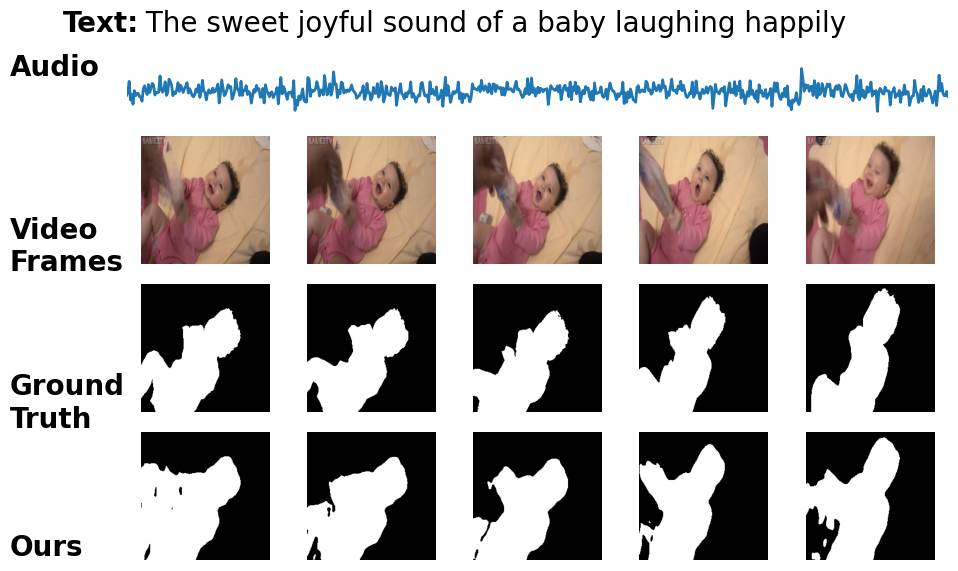}
    }
    \caption{Visualization of audio-to-text generalization on AVS-S4  video segmentation task – Baby Laughing}
    \label{fig:A2T_baby}
\end{figure}

\begin{figure}
\resizebox{0.9\columnwidth}{!}{%
    \centering
    \includegraphics[width=1\linewidth]{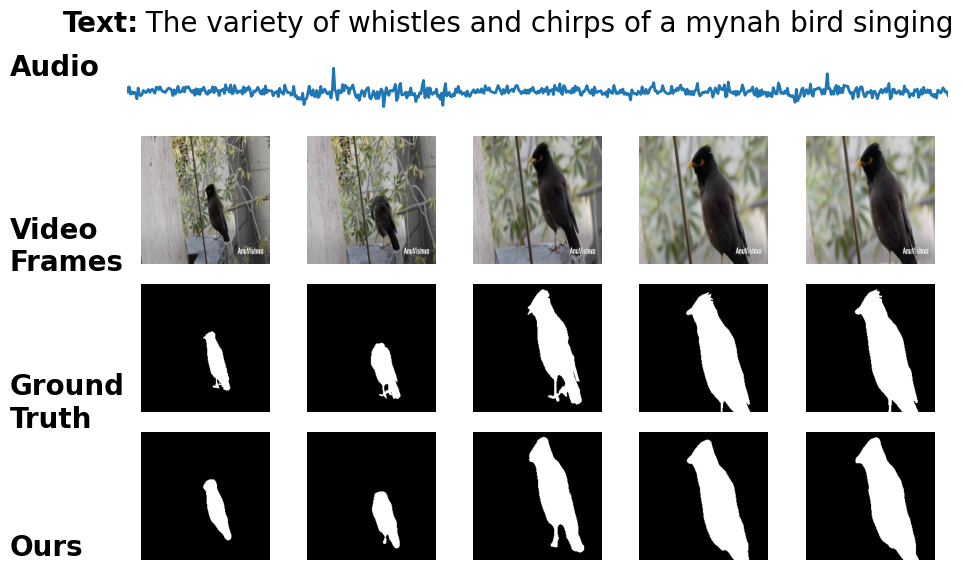}
    }
    \caption{Visualization of audio-to-text generalization on AVS-S4  video segmentation task – Mynah Bird Signing}
    \label{fig:A2T_mynah}
\end{figure}
\noindent
We visualize cross-modal transfer on AVSBench-S4~\cite{AVS} using our frozen encoders and trimodal codes, training a downstream query-based segmentation head that follows the AVS architecture~\cite{AVS}. Trained with \emph{text} queries and evaluated with \emph{audio} inputs (T $\!\rightarrow$ A), the model accurately segments the visual sources corresponding to the sound, as seen in Fig.~\ref{fig:T2A_piano} (piano playing) and Fig.~\ref{fig:T2A_coyote} (coyote howling). Conversely, trained with \emph{audio} queries and evaluated with \emph{text} (A $\!\rightarrow$ T), the model produces clean masks aligned with the queried semantics, e.g., Fig.~\ref{fig:A2T_baby} (baby laughing) and Fig.~\ref{fig:A2T_mynah} (mynah bird singing). This highlights robust cross-modal generalization for video segmentation. The performance advantage is attributed to our tri-modal codes, which preserve richer visual semantics through our Cascading Semantic Alignment module.

\section{CSA Closed-Form Weights}\label{app:csa-closed}

\noindent\textbf{Summary:} 
% CSA aligns modality-specific centroids at index $k$ by (i) forming a trimodal anchor $\mathbf{c}^0(k)$ via the equation in \eqref{eq:cm_center}, and (ii) applying a sequential T $\!\rightarrow\!$ A $\!\rightarrow\!$ V update in \eqref{eq:cascade_short}. 
% Using fixed, non-negative coefficients keeps all updates inside the convex hull of $\{\mathbf{e}_t^0,\mathbf{e}_a^0,\mathbf{e}_v^0\}$, creating progressive centroids and stabilizing learning while avoiding codebook collapse.
CSA aligns modality-specific centroids at index $k$ by forming a trimodal anchor $\mathbf{c}^0(k)$ (\eqref{eq:cm_center}) and applying a sequential T$\!\rightarrow\!$A$\!\rightarrow\!$V update (\eqref{eq:cascade_short}). Fixed non-negative coefficients keep updates inside the convex hull of $\{\mathbf{e}_t^0,\mathbf{e}_a^0,\mathbf{e}_v^0\}$, creating progressive centroids, stabilizing learning and preventing codebook collapse.

\noindent\textbf{Closed-form coefficients:}
Unrolling \eqref{eq:cascade_short} yields
\begin{equation}
\label{eq:csa_closed}
\begin{aligned}
\mathbf{e}_t^1(k) &= \tfrac{1}{3}\mathbf{e}_t^0(k)+\tfrac{1}{3}\mathbf{e}_a^0(k)+\tfrac{1}{3}\mathbf{e}_v^0(k),\\[-2pt]
\mathbf{e}_a^1(k) &= \tfrac{1}{9}\mathbf{e}_t^0(k)+\tfrac{4}{9}\mathbf{e}_a^0(k)+\tfrac{4}{9}\mathbf{e}_v^0(k),\\[-2pt]
\mathbf{e}_v^1(k) &= \tfrac{4}{27}\mathbf{e}_t^0(k)+\tfrac{7}{27}\mathbf{e}_a^0(k)+\tfrac{16}{27}\mathbf{e}_v^0(k).
\end{aligned}
\end{equation}
The video centroid receives the largest self-weight ($16/27$), so it retains the most modality-specific detail while still moving toward the multimodal consensus.

\noindent\textbf{Effect:}
Together, temporally aligned EMA aggregation (DTA), the CSA cascade above, and the commitment loss (Eq.~\eqref{eq:commitment})  yield \emph{distinct multimodal semantic spheres}, as visualized in Fig.~\ref{fig:sub:seq_ema_1} and Fig.~\ref{fig:cascade}.

\clearpage

\end{document}